\documentclass{article}

\usepackage[margin=1in]{geometry}
\usepackage{authblk}
\usepackage{graphicx}
\usepackage[utf8]{inputenc} %
\usepackage[T1]{fontenc}    %
\usepackage[pdfencoding=auto, psdextra,hidelinks]{hyperref}       %
\usepackage{url}            %
\usepackage{booktabs}       %
\usepackage{amsfonts}       %
\usepackage{nicefrac}       %
\usepackage{microtype}      %
\usepackage{xcolor}         %
\usepackage{amsmath}
\usepackage{amssymb}
\usepackage{amsthm}
\usepackage{mathtools}
\usepackage{subcaption}
\usepackage[capitalize,noabbrev]{cleveref}
\usepackage[numbers]{natbib}
\usepackage{multirow}
\usepackage{wrapfig}

\usepackage{comment}
\usepackage{enumitem}

\theoremstyle{plain}
\newtheorem{theorem}{Theorem}[section]
\newtheorem{proposition}[theorem]{Proposition}
\newtheorem{lemma}[theorem]{Lemma}

\theoremstyle{definition}
\newtheorem{definition}[theorem]{Definition}
\newtheorem{assumption}[theorem]{Assumption}
\theoremstyle{remark}

\crefname{proposition}{proposition}{propositions}
\Crefname{proposition}{Proposition}{Propositions}

\newcommand{\reals}{\ensuremath{\mathbb{R}}}

\DeclareMathOperator*{\Ex}{\mathbb{E}}
\newcommand{\PP}{\mathbb{P}}

\newcommand{\diag}[1]{\mathrm{diag}\left(#1\right)}

\newcommand{\argmax}{\mathop{\mathrm{argmax}}}

\newcommand{\norm}[1]{\left\lVert#1\right\rVert}

\newcommand{\dataset}{\ensuremath{\mathcal{D}}}
\newcommand{\basedist}{\ensuremath{\PP_{0}}}
\newcommand{\colldist}{\ensuremath{\PP^*}}

\newcommand{\mixdist}{\ensuremath{\PP_{\strength}}}
\newcommand{\collfn}{\ensuremath{g}}

\newcommand{\opti}{\ensuremath{\epsilon}}

\newcommand{\strength}{\ensuremath{\alpha}}

\newcommand{\classifier}{\ensuremath{h}}

\newcommand{\sensitivity}{\ensuremath{\tau}}
\newcommand{\Xtest}{\ensuremath{X_\text{min}}}

\newcommand{\compas}{COMPAS}
\newcommand{\adult}{Adult}
\newcommand{\hsls}{HSLS}
\newcommand{\acsincome}{ACS-Income}
\newcommand{\birdspca}{Waterbirds-PCA}
\newcommand{\birdsfull}{Waterbirds-Full}
\newcommand{\civilpca}{CivilComments-PCA}
\newcommand{\civilfull}{CivilComments-Full}
\newcommand{\gmmfour}{\ensuremath{\PP_\text{4GMM}}}

\newcommand{\numflips}{\ensuremath{M}}
\newcommand{\bylabel}{RB-label}
\newcommand{\bydistance}{RB-dist}
\newcommand{\byprobability}{RB-prob}
\newcommand{\lblerror}{\ensuremath{\rho}}	

\newcommand{\mytitle}{{Minority} {Collective} {Action} {for} {User}-{Side} {Fairness}}

\title{\mytitle{}}

\author[1]{Omri Ben-Dov}
\author[1]{Samira Samadi}
\author[2]{Amartya Sanyal}
\author[3]{Alexandru \c{T}ifrea}

\affil[1]{Max Planck Institute for Intelligent Systems, Tübingen AI Center, Tübingen, Germany}
\affil[2]{Department of Computer Science, University of Copenhagen}
\affil[3]{ETH Zurich}

\date{}

\begin{document}

\maketitle

\begin{abstract}
Machine learning models often preserve biases present in training data, leading to unfair treatment of certain minority groups. Despite an array of existing firm-side bias-mitigation techniques, these methods typically incur utility costs and require organizational buy-in. Recognizing that many models rely on user-contributed data, end-users can induce fairness through the framework of Algorithmic Collective Action. In this setting, a coordinated minority group strategically relabels its own data to enhance fairness without altering the firm's training process. We design practical, model-agnostic, minority-only collective-action methods and validate them on real-world datasets. Our findings show that a subgroup of the minority can substantially reduce unfairness with little impact on overall prediction error.
\end{abstract}

    \section{Introduction}
As machine learning (ML) tools become increasingly accessible, more firms deploy them for decision-making.
However, ML models often perpetuate societal biases present in their training data, leading to unfair outcomes across demographic groups~\citep{barocasBigDatasDisparate2016}.
Moreover, most fairness-preserving learning algorithms incur a non-negligible cost in accuracy or computational resources~\citep{menonCostFairnessBinary2018,zhaoInherentTradeoffsLearning2022,dehdashtianUtilityfairnessTradeoffsHow2024,sadeghiCharacterizingTradeoffInvariant2022}, which can discourage practical adoption.
Since firms control the ML pipeline, end-users lack access to these algorithms and cannot directly enforce fair treatment.
Yet, affected users routinely generate and share data, through clicks, ratings, or other contributions, that is used to train the firm's models.
Such cases, where users collaborate to influence what firms learn, are not uncommon and are well-documented~\citep{siggDeclineNowCombinatorial2025}.
Consequently, this paper asks whether underrepresented minority groups can collaboratively alter the data they share to steer learned models toward fairer behavior, even without access to the firm’s training pipeline.

For example, consider a human resources company that fills vacancies and trains ML models on resumes to predict candidates' skills.
While the majority may have more formal education and college degrees, disadvantaged groups may have informal training or internships.
As a result, the ML model may fail to assign the correct skills to minority candidates because they lack formal education, despite their practical experience.
Minority members can react to this injustice by collectively submitting their resumes while reframing their reported skills, such as sales or management.
\cref{ap:motivation} describes other examples where such \emph{collective action} is applicable.
This idea is reminiscent of \emph{pre-processing} fairness techniques~\citep{kamiranClassifyingDiscriminating2009,luongKNNImplementationSituation2011,zemelLearningFairRepresentations2013,madrasLearningAdversariallyFair2018}, which modify the data before model training.
Unlike these prior approaches, which assume centralized control over the data, we consider the setting of \emph{algorithmic collective action} (ACA)~\citep{hardtAlgorithmicCollectiveAction2023,ben-dovRoleLearningAlgorithms2024,baumannAlgorithmicCollectiveAction2024,siggDeclineNowCombinatorial2025,gauthierStatisticalCollusionCollectives2025}, in which a small group of users strategically modifies their own data to influence the correlations learned by the model.

\begin{figure*}
    \centering
    \begin{subfigure}{0.45\linewidth}
        \centering
        \caption{Before collective action}
        \includegraphics[width=\linewidth]{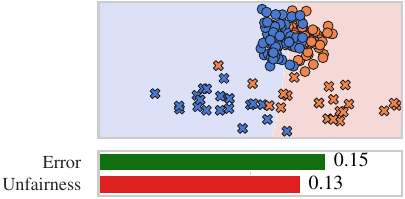}
    \end{subfigure}
    \hfill
    \begin{subfigure}{0.45\linewidth}
        \centering
        \caption{After collective action}
        \includegraphics[width=\linewidth]{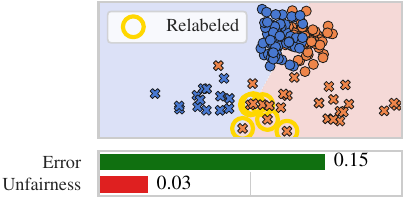}
    \end{subfigure}
    \caption{
    Minority-only collective action can substantially improve fairness.
    When only 6 minority members change their labels, the EqOd violation (\cref{eq:EqOd}) of logistic regression drops by over $75\%$ with a negligible rise in prediction error~(\cref{eq:error_rate}).
    Circles represent the majority and crosses represent the minority.
    }
    \label{fig:teaser}
\end{figure*}

\subsection{Our contributions}
\paragraph{Shifting focus from firm-side fairness methods to the user side.}
After formally introducing fairness in ML and ACA in \cref{sec:problem}, we adapt the \emph{erasure strategy} from \citet{hardtAlgorithmicCollectiveAction2023} to reduce predictive correlation between group membership and the target label by relabeling minority samples. By erasing a counterfactual signal, the success metric of the collective aligns with the definition of counterfactual fairness, which under certain conditions can translate to group fairness.

\paragraph{Designing minority-only collective action with counterfactual estimation.}
The erasure strategy faces two major challenges in our setting.
First, the collective is limited to only minority members, since minority members are more motivated to join collective action~\citep{saleemWhenHowNegative2021,begenyPowerIngroupPromoting2022} and can be efficiently mobilized~\citep{mcadamPoliticalProcessDevelopment1999,michelsonMeetingChallengeLatino2005}, while majority-group users may be less inclined to disrupt the status quo.
Second, the optimal strategy requires counterfactual labels which are generally intractable.
To overcome these obstacles, in \cref{sec:method} we propose a strategy in which only a specific number of candidates flip their labels.
The candidates are ranked by one of three simple methods that estimate how likely a candidate is to have a positive counterfactual label.
To test this strategy, in \cref{chapter:results} we conduct experiments on common real-world fairness datasets. As we cannot evaluate counterfactual fairness, we instead measure group fairness metrics as surrogates.
These experiments show that the proposed strategy substantially decreases EqOd violation with only a small impact on prediction error, as illustrated in \cref{fig:teaser}.

\paragraph{Limitations of minority-only collectives.}
\cref{ssec:minority_restrictions} investigates the fundamental limitations of minority-only collectives, such as the impossibility of reaching zero EqOd violation, and provides theoretical results showing that learned representations and accurate counterfactual approximation methods can improve on the basic strategy we propose.

    \section{Collective Action for Fairness}
\label{sec:problem}

To establish the connection between fairness and ACA, \cref{ssec:group_fairness} first defines the problem setting and how unfairness can be measured.
Then, \cref{ssec:collective} describes the theoretical framework of ACA and how it can be utilized to mitigate bias.
Finally, \cref{ssec:counterfactual_fairness} formally relates ACA to group fairness metrics through counterfactual fairness.

\subsection{Group fairness for classification}
\label{ssec:group_fairness}

In our setting, a firm uses ML to predict a binary label \(y\in\{0,1\}\).
The firm collects data from its users, forming a dataset \(\dataset{=}\left\{ \left(x_{i},a_{i},y_{i}\right)\right\} _{i=1}^{n}\), where \(x_{i}\in\reals^{m}\) denotes user \(i\)'s feature vector,~\(a_{i}\in\{0,1\}\) is a sensitive attribute indicating binary group membership (\(a_{i}=0\) for the majority group, \(a_{i}=1\) for the minority), and \(y_{i}\in\{0,1\}\) is the true label.
We assume the users are drawn independently and identically distributed (i.i.d.) from a distribution \basedist{} over \(\reals^{m}{\times}\left\{ 0,1\right\} {\times}\left\{ 0,1\right\} \).
The firm trains a classifier $\classifier:\reals^m\to\{0,1\}$ to minimize the prediction error, defined as
\begin{equation} \label{eq:error_rate}
    \mathrm{Error}\left(\classifier\right)=\PP\left[\classifier\left(x\right)\neq y\right].
\end{equation}
To do so, the firm minimizes the empirical error on \dataset{} via Empirical Risk Minimization (ERM).

In the group-fairness paradigm, the sensitive attribute \(a\in\{0,1\}\) partitions the data into subgroups, and fairness criteria seek to ensure similar outcomes across these groups.
Common metrics include statistical parity (SP)~\citep{caldersBuildingClassifiersIndependency2009,dworkFairnessAwareness2012} and equalized odds (EqOd)~\citep{hardtEqualityOpportunitySupervised2016}.
In this work, we focus primarily on violations of EqOd, formally defined as
\begin{equation}\label{eq:EqOd}
    \text{EqOd}\left(\classifier\right)=\frac{1}{2}\sum_{z=0,1}\left|\PP\left[\classifier\left(x\right)=1|a=1,y=z\right]-\PP\left[\classifier\left(x\right)=1|a=0,y=z\right]\right|\text{,}
\end{equation}
which measures the differences between true positive and false positive rates.
\cref{ap:fairness_metrics} provides formal definitions and further discussion of these metrics.

ERM-trained models tend to achieve low predictive error, but often at the cost of SP and EqOd violations~\citep{menonCostFairnessBinary2018,zhaoInherentTradeoffsLearning2022,bardenhagenBoostingWorstgroupAccuracy2021,sanyalHowUnfairPrivate2022}.
Despite significant progress in fairness research, most solutions have traditionally focused on \emph{firm-side} solutions: pre-processing the dataset, in-processing modifications to the training algorithm, or post-processing the classifier's predictions.
These approaches almost always incur errors or additional pipeline complexity, discouraging firms from deploying them in practice.

While most prior work has focused on firm-side solutions, this work shifts the focus to \emph{user-side} methods that do not require the firm's participation.
Since users generate the training data, they can collectively influence the learned model by strategically modifying their own behavior.
For instance, consider a digital platform that recommends content to a user based on a classifier that predicts engagement labels \(y_{i}\in\left\{ \text{will engage},\text{ will not engage}\right\} \).
The classifier, trained on historical user interactions, may unintentionally rely on group membership rather than individual preferences when making recommendations for minority members.
In response, users can coordinate to alter interaction patterns, such as clicking on or avoiding certain items.
This ACA affects the dataset in a way that steers the learned classifier toward fairer outcomes, and is generally studied in the field of algorithmic collective action~\citep{hardtAlgorithmicCollectiveAction2023}.

\subsection{Algorithmic collective action} \label{ssec:collective}

In social sciences, \emph{collective action} refers to the coordinated efforts of individuals working together to pursue a shared goal~\citep{olsonCollectiveAction1989,marwellCriticalMassCollective1993}.
\citet{hardtAlgorithmicCollectiveAction2023} adapt this notion to ML, proposing that a group of users, termed a collective, can strategically modify their data to align the behavior of a trained classifier \classifier{} with the collective's goals.
In this formulation, the training distribution is a mixture distribution \(\dataset\sim\mixdist  = \strength\colldist + (1-\strength)\basedist\), where \colldist{} and \basedist{} are the collective and base distributions, and \(\strength\in\left[0,1\right]\) denotes the proportion of the collective.%

\paragraph{Relation to fair representation learning.}
With user agency over the data, one possible form of ACA for fairness is to modify their features to increase correlation with the label \(y=1\).
An analogous firm-side approach is fair representation learning (FRL), which learns a transformation from the input space to a representation space such that ERM leads to a classifier that is both accurate and fair~\citep{zemelLearningFairRepresentations2013,jovanovicFAREProvablyFair2023}.
However, a hindrance of FRL in the context of ACA is that the transformation must be applied consistently at inference time, requiring active cooperation from each minority member to transform their features.
In contrast, our setting assumes users have control only over the labels and cannot intervene in other parts of the machine learning pipeline.

\paragraph{Erasing a signal.} Suppose the collective seeks a classifier that is invariant under a transformation~\(\collfn: \reals^{m} \to \reals^{m}\) applied to the features.
The success of the collective is 
\begin{equation} \label{eq:success_def}
S\left(\strength\right)=\basedist\left[\classifier\left(\collfn\left(x\right)\right)=\classifier\left(x\right)\right],
\end{equation}
the probability, under the base distribution, that the classifier's prediction is unchanged after applying \collfn{} to the features.
In words, the collective's goal is to \textit{erase the signal} \collfn{}: to ensure the classifier behaves identically regardless of whether \collfn{} is applied.
Intuitively, if \collfn{} embeds a pattern correlated with group membership (i.e., minority or majority), then achieving invariance under \collfn{} promotes fairness by reducing the classifier's dependence on group-identifying information.

To achieve signal erasure, \citet{hardtAlgorithmicCollectiveAction2023} propose that the collective relabels itself with the most likely label under the transformation \collfn{}.
Formally, the strategy is defined as
\begin{equation} \label{eq:erasure-strategy}
    x,y\rightarrow x,\argmax_{y^{\prime}\in\left\{ 0,1\right\} }\basedist\left(y^{\prime}|g\left(x\right)\right).
\end{equation}
Since this strategy leaves the features unchanged, it is well-suited for settings where the minority is limited to modifying only their labels, such as ours.
For \(\opti{}\)-optimal Bayes classifiers (\cref{def:clf-opt} in \cref{ap:clf-opt}), \citet{hardtAlgorithmicCollectiveAction2023} prove the following lower bound for its success
\begin{equation} \label{eq:success_bound}
    S\left(\strength\right)\geq1-\frac{2\left(1-\strength\right)}{\strength}\cdot\sensitivity-\frac{\opti}{\left(1-\opti\right)\strength},
\end{equation}
where \(\displaystyle \sensitivity=\Ex_{x\sim\basedist}\left[\max_{y^{\prime}\in\left\{ 0,1\right\} }\left|\basedist\left(y^{\prime}|x\right)-\basedist\left(y^{\prime}|g\left(x\right)\right)\right|\right]\) captures the sensitivity of \(y\) under \collfn{}.

Note that the strategy in \cref{eq:erasure-strategy} may require some majority members to relabel themselves with the label \(y=0\).
Such a change might deter them from participating in the collective action, either because majority members are unwilling to give up their advantage or prefer to maintain the status quo.
To avoid this conflict, we restrict the collective to include only minority members.
We discuss the implications of this restriction in \cref{ssec:minority_restrictions}.

\subsection{Counterfactual fairness}
\label{ssec:counterfactual_fairness}
The concept of \emph{counterfactual fairness} (CF)~\citep{kusnerCounterfactualFairness2017,gargCounterfactualFairnessText2019,wuCounterfactualFairnessUnidentification2019} connects signal-erasure success to group fairness.
To introduce this idea, assume that a sample \(x\) is generated by a causal model, in which the group membership \(A\) is a causal parent.
Then a classifier \classifier{} is counterfactually fair if its predictions are invariant to interventions on the group membership, i.e., \(\classifier\left(x\right)=\classifier\left(x_{A\leftarrow a^{\prime}}\right)\) for any \(a^{\prime}\), where \(x_{U\leftarrow u}\) denotes an intervention on a causal parent \(U\) of a sample \(x\).
In certain causal contexts, CF implies or aligns with group fairness criteria such as SP or EqOd~\citep{anthisCausalContextConnects2023}.
Therefore, if ACA induces a counterfactually fair classifier, it may also induce a fair classifier under SP or EqOd.
As our focus is on fairness for the minority, we relax the original definition of CF~\citep{kusnerCounterfactualFairness2017}.
\begin{definition} \label{def:onesidedcounterfactual_fair}
    A classifier \(h\) is \textbf{minority-focused counterfactually fair} if, under the causal distribution induced by \basedist{},
    \begin{equation}
    \basedist\left(h\left(x_{A\leftarrow 1}\right)=y|X=x,A=1\right)=\basedist\left(h\left(x_{A\leftarrow a^{\prime}}
    \right)=y|X=x,A=1\right)\text{,}
    \end{equation}
    for all \(y\) and for any value \(a^{\prime}\) attainable by \(A\).
\end{definition}
By this definition, changing the group membership of a minority
individual, in a counterfactual sense, has no effect on the
classifier's prediction.  ACA can theoretically enforce such fairness by applying the erasure strategy from \cref{eq:erasure-strategy} with
the counterfactual signal \(\collfn\left(x\right)=x_{A\leftarrow0}\),
which replaces a minority individual with its majority-group
counterfactual.  This ACA aligns the signal erasure success from
\cref{eq:success_def} with minority-focused counterfactual fairness
from \cref{def:onesidedcounterfactual_fair}.  The following informal
proposition, proved in \cref{ap:corl1_proof}, formalizes this
alignment.

\begin{proposition}[Informal; see Proposition~\ref{prop:success_cf}]\label{prop:success_cf_inf}
    For a deterministic Bayes classifier, perfect success of the
    minority collective is equivalent to~(almost-sure) minority-focused counterfactual fairness.
\end{proposition}

This result directly connects ACA theory to fairness.
Thus, perfect success of the collective is equivalent to achieving minority-focused counterfactual fairness.
 
    \section{Minority Collective Action With Approximated Counterfactuals}
\label{sec:method}

\begin{figure*}
    \centering
    \begin{subfigure}{0.3\linewidth}
        \centering
        \caption{\bylabel{}} \label{fig:bylabel}
        \includegraphics[width=\linewidth]{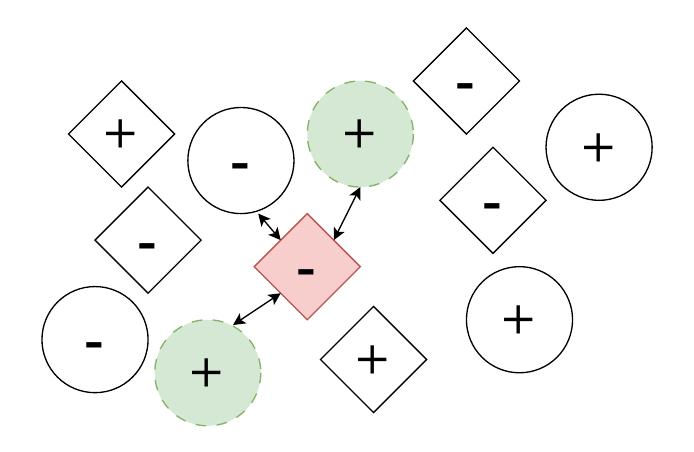}
    \end{subfigure}
    \begin{subfigure}{0.3\linewidth}
        \centering
        \caption{\bydistance{}} \label{fig:bydistance}
        \includegraphics[width=\linewidth]{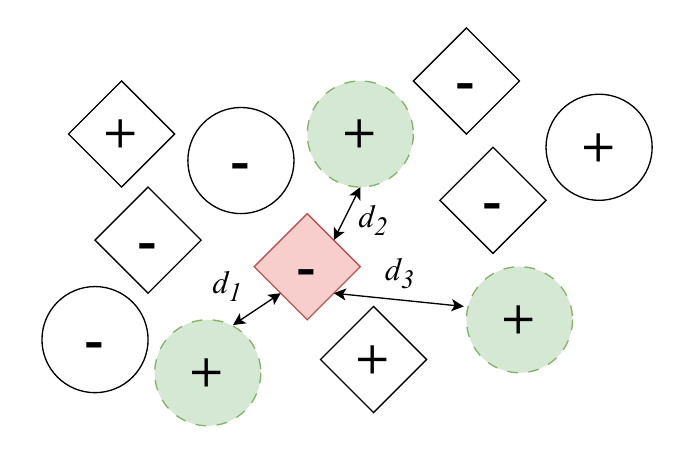}
    \end{subfigure}

    \caption{
    Visualization of KNN scoring methods in \cref{sec:method} with \(k{=}3\).
    Squares represent the minority and circles the majority, marked with a positive ``\(+\)'' or a negative ``\(-\)'' label.
    (\textbf{a}) \textit{\bylabel{}}: Two of the nearest majority neighbors have a positive label, resulting in the score \(s{=}2\).
    (\textbf{b}) \textit{\bydistance{}:} The average distance to the nearest positive majority neighbors results in the score~\(s{=}-\left(d_{1} {+} d_{2} {+} d_{3}\right)/3\).}
    \label{fig:vis-method}
\end{figure*}

This section describes how a minority collective can approximate a signal-erasure strategy to promote fairness in practice.
While the theory of signal erasure has been studied before~\citep{hardtAlgorithmicCollectiveAction2023,gauthierStatisticalCollusionCollectives2025}, prior work lacks empirical evaluation.
In this paper, we present the first practical algorithm for signal erasure and provide experimental results in \cref{chapter:results}.
As discussed in \cref{ssec:counterfactual_fairness}, a suitable signal to erase is \(\collfn\left(x\right) = x_{A\leftarrow0}\), where each collective member relabels themselves according to \cref{eq:erasure-strategy}.

However, end-users lack access to the causal model and cannot compute the counterfactual labels directly.
To address this limitation, we propose to assign each collective member \(i\) a score \(s_{i}\), which serves as a proxy for the likelihood that they would receive the label \(y = 1\) if they belonged to the majority.
Given a budget of \numflips{} label flips, the collective selects the \numflips{} members with the highest scores; these individuals flip their labels from \(y=0\) to \(y=1\).
The budget \numflips{} controls the accuracy--fairness tradeoff: a higher budget typically improves fairness but increases error.%

We suggest three scoring functions, each capturing a different notion of similarity to majority users:
\begin{enumerate}[leftmargin=1mm]
    \item \textbf{Rank by probability (\byprobability{}):} Train a regressor \(f:\mathbb{R}^{m}\to\mathbb{R}\) exclusively on majority data ($a=0$) to estimate the probability \(\PP\left(Y=1|X=x,A=0\right)\) of having the label \(y=1\) under the majority distribution.
    Each collective member \(i\) receives a score based on the model's prediction:
    \begin{equation}
        s_{i}=f\left(x_{i}\right).
    \end{equation}
    \item \textbf{Rank by label (\bylabel{}):} 
    For each collective member \(i\), identify the set \(K_{i}\) of their \(k\)-nearest majority neighbors using Euclidean distance.
    The score is the number of neighbors with the label \(y=1\):
    \begin{equation}
        s_{i}=\sum_{j\in K_{i}}\mathbf{1}\left\{ y_{j}=1\right\} .
    \end{equation}
    \item \textbf{Rank by distance (\bydistance{}):} Restrict the set of neighbors \(K_{i}\) to only majority users with the label $y=1$.
    The score is the negative mean Euclidean distance to these neighbors:
    \begin{equation}
        s_{i}=-\frac{1}{k}\sum_{j\in K_{i}}\|x_{i}-x_{j}\|_{2}.
    \end{equation}
\end{enumerate}

Intuitively, \byprobability{} acts as a naive counterfactual estimator by training only on majority data.
\bylabel{} assumes that a sample with more positive-labeled majority neighbors is more likely to have a positive counterfactual label, and \bydistance{} assumes that proximity to positive-labeled majority samples is evidence for a positive counterfactual label.
Note that \bylabel{} and \bydistance{} are \(k\)-nearest neighbors (KNN) based methods, which are common in counterfactual estimation~\citep{stuartMatchingMethodsCausal2010,liMatchingDimensionalityReduction2016,vermaCounterfactualExplanationsAlgorithmic2024}.

    \section{Experimental Results} \label{chapter:results}

This section evaluates the performance of our methods.
However, since we cannot compute the counterfactual fairness gap in \cref{def:onesidedcounterfactual_fair}, we instead measure surrogate group fairness metrics.
The main text presents results for EqOd, while \cref{ssec:expanded_results} presents results for both EqOd and SP, along with additional datasets.
We compare the three methods, \bylabel{}, \bydistance{}, \byprobability{}, against a random baseline that flips \(y=0\) labels to \(y=1\) for \numflips{} randomly selected collective members.
We conducted experiments on the tabular datasets \compas{}~\citep{mattuHowWeAnalyzed2016}, \adult{}~\citep{beckerAdult1996}, \hsls{}~\citep{jeongFairnessImputationDecision2022}, and \acsincome{}~\citep{dingRetiringAdultNew2021}, the image dataset Waterbirds~\citep{sagawaDistributionallyRobustNeural2020}, and the text dataset CivilComments~\citep{borkanNuancedMetricsMeasuring2019}.
For Waterbirds, we use features extracted from a pre-trained \textit{ResNet-18}~(denoted \birdsfull{}), and for CivilComments, we use features extracted from Hugging Face's pre-trained \textit{bert-base-uncased} model~(denoted \civilfull{}).
In addition to the complete features of Waterbirds and CivilComments, we also include experiments on the PCA features, with 85 components for Waterbirds~(denoted \birdspca{}) and 100 components for CivilComments~(denoted \civilpca{}).
Details on the datasets and preprocessing are provided in \cref{ap:datasets}.

All reported metrics are computed on a fixed test set without any ACA and averaged over 10 independent runs for each method described in \cref{sec:method}.
In each run, we randomly selected a minority collective to apply the method.
For the KNN-based methods, we tuned the neighborhood size $k$ using a \(15\%\) validation split from the train set, optimizing for EqOd and SP.
Finally, we trained a gradient-boosted decision tree on each modified train set.
More technical details are given in \cref{ap:training}.

\begin{figure*}
    \centering
    \begin{subfigure}{0.31\linewidth}
        \centering
        \caption{\compas{}}
        \includegraphics[width=\linewidth]{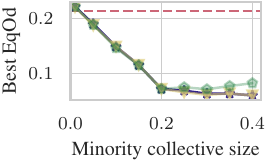}
    \end{subfigure}
    \hfill
    \begin{subfigure}{0.31\linewidth}
        \centering
        \caption{\acsincome{}}
        \includegraphics[width=\linewidth]{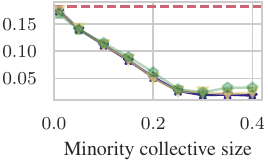}
    \end{subfigure}
    \hfill
    \begin{subfigure}{0.31\linewidth}
        \centering
        \caption{\birdsfull{}}
        \includegraphics[width=\linewidth]{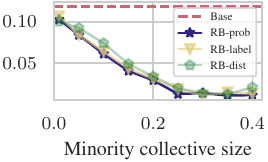}
    \end{subfigure}
    
    \caption{
    The lowest EqOd violation a collective can achieve decreases as the collective size increases, up to a certain point.
    Each point is the mean of 10 runs, with the standard deviation being smaller than the markers.
    Across the datasets we experimented on, the lowest EqOd violation stabilizes around \(\strength=0.3\).
    Additional results are presented in \cref{fig:sp_effect_alpha} in the appendix.
    }
    \label{fig:compare_alpha}
\end{figure*}

\paragraph{Importance of collective size}
While the number of label flips \numflips{} is the primary factor for balancing between accuracy and fairness, the size of the collective, \strength{}, also plays a role.
In addition to bounding the possible number of flips, increasing \strength{} also expands the candidate pool from which the most effective labels to flip can be selected.
To measure this effect, our experiments include a range of \strength{} values, each tested with multiple values of \numflips{}.
For each \strength{}, we define the best achievable EqOd as the minimum EqOd across all tested values of \numflips{}.
As shown in \cref{fig:compare_alpha}, increasing \strength{} decreases the lowest achievable EqOd violation until it saturates around \(\strength=0.3\).
We fix this value for all remaining experiments.

\begin{figure*}
    \centering
    \begin{subfigure}{0.32\linewidth}
        \centering
        \caption{\compas{}}
        \includegraphics[width=\linewidth]{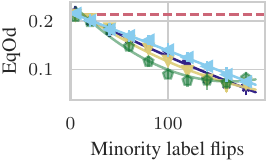}
        \label{fig:flip_sub_compas}
    \end{subfigure}
    \hfill
    \begin{subfigure}{0.32\linewidth}
        \centering
        \caption{\acsincome{}}
        \includegraphics[width=\linewidth]{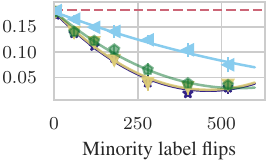}
        \label{fig:flip_sub_income}
    \end{subfigure}
    \hfill
    \begin{subfigure}{0.32\linewidth}
        \centering
        \caption{\birdsfull{}}
        \includegraphics[width=\linewidth]{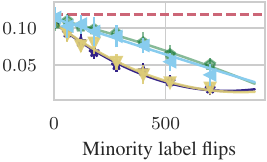}
        \label{fig:flip_sub_birds}
    \end{subfigure}
    \includegraphics{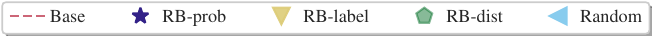}
    
    \caption{
    Our proposed methods are generally more efficient than randomly flipping labels, requiring fewer label flips to attain the same EqOd violation level.
    Each marker is the mean of 10 random runs with a specific number of label flips. Error bars show the standard deviation.
    The dashed line shows the mean EqOd for a classifier trained on the dataset without collective action.
    }
    \label{fig:compare_flips}
\end{figure*}

\paragraph{Flipping cost} \label{sec:cost_flips}
Since each method scores candidates differently, they may also vary in efficiency, that is, the number of label flips required to achieve a given level of EqOd violation.
To evaluate efficiency, \cref{fig:compare_flips} plots EqOd as a function of number of label flips \numflips{}, where lower curves indicate more efficient methods.
The random baseline consistently yields the worst EqOd across all values of \numflips{}, highlighting the value of informed relabeling algorithms.
However, no single method dominates the others in all settings:
while \byprobability{} and \bylabel{} often outperform the other methods, \bydistance{} can surpass them in specific cases (e.g., \cref{fig:flip_sub_compas}), or perform comparably to the random baseline in others (\cref{fig:flip_sub_birds}).
These results suggest that a well-chosen scoring function enables the collective to achieve a desired level of EqOd violation with fewer label flips, reducing the ``cost'' of ACA and mitigating the accuracy loss from excessive relabeling.

Interestingly, beyond a certain number of flips, EqOd begins to increase, indicating that excessive flipping can shift unfairness from the minority to the majority.
This upturn reflects the fundamental limits of minority ACA for fairness, a point we elaborate on in \cref{ssec:minority_restrictions}.

\begin{figure*}
    \centering
    \begin{subfigure}{0.31\linewidth}
        \centering
        \caption{\compas{}, \byprobability{}}
        \includegraphics[width=\linewidth]{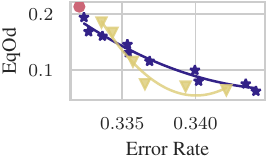}
    \end{subfigure}
    \hfill
    \begin{subfigure}{0.31\linewidth}
        \centering
        \caption{\acsincome{}, \byprobability{}}
        \includegraphics[width=\linewidth]{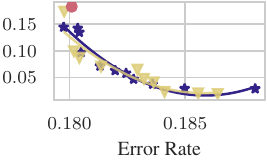}
    \end{subfigure}
    \hfill
    \begin{subfigure}{0.31\linewidth}
        \centering
        \caption{\birdsfull{}, \byprobability{}}
        \includegraphics[width=\linewidth]{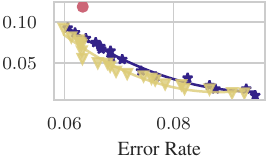}
    \end{subfigure}
    \includegraphics{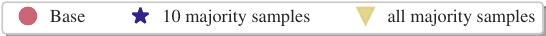}
    
    \caption{
    Limiting the collective's knowledge of the majority does not significantly harm the Pareto front.
    Each point is the mean of 10 runs and the curves are fitted to guide the eye.
    }
    \label{fig:compare_knowledge}
\end{figure*}

\paragraph{Partial knowledge of the majority} \label{ssec:partial}
In the previous experiments, we assumed that the collective has full access to the majority data. %
Here we investigate the performance of our methods when this knowledge is limited.
To visualize the fairness-error tradeoff, we measure the error and EqOd for a range of label flips, yielding a set of pairs (Error, EqOd).
This set forms a Pareto front, representing the tradeoff.
A Pareto front is said to \emph{dominate} another if it lies entirely to the left (lower error) and below (lower unfairness) of the other.
The Pareto fronts in \cref{fig:compare_knowledge} show that a collective employing \byprobability{}, when restricted to only 10 majority members, performs similarly to a collective with full knowledge.
While the Pareto fronts remain similar, limited majority knowledge can increase the number of required flips.
This is evident when comparing to the zero-knowledge scenario, designated as random in \cref{fig:compare_flips}.
This finding implies that the fewer flips the collective is allowed, the more important it is to have access to the majority data.

    \section{Limitations of Minority Collective Action} \label{ssec:minority_restrictions}

\begin{wrapfigure}{r}{0.32\textwidth}
\vspace{-35pt}
  \begin{center}
    \includegraphics[width=0.25\textwidth]{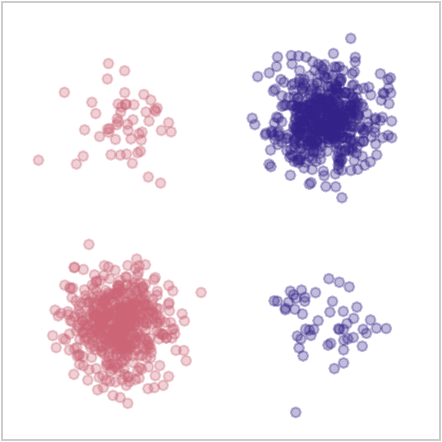}
  \end{center}
  \caption{The distribution $\PP_\text{4GMM}$ used in \cref{corol:impossible_fairness}.
  The color signifies the label, and the density shows the group membership.}
  \label{fig:example_dist}
  \vspace{-20pt}
\end{wrapfigure}

Previous work on ACA assumes that the collective is uniformly sampled from the distribution \basedist{} and that the collective has a perfect oracle for the conditional distribution \(\basedist{}\left(Y|X\right)\).
Yet, our method restricts collective participation to minority members and approximates this conditional distribution.
These differences introduce limitations to the existing theory, which we analyze
in this section.

\paragraph{Collective restricted to the minority.}
\label{sec:minority_restriction}

As mentioned above, we focus on minority collectives, unlike prior work.
This restriction captures scenarios in which majority members lack incentives to support changes that would benefit the minority and instead prefer to preserve the status quo.
Naturally, this limitation reduces the collective's impact.
Consider a binary classification task on the two-dimensional 4-Gaussian mixture model \gmmfour{} where each Gaussian belongs to a distinct combination of label and group membership, as illustrated in \cref{fig:example_dist}.
Each label consists of a large majority subgroup and a significantly smaller minority subgroup.
We can then state the following informal result about the EqOd violation of ERM.

\begin{proposition}[Informal] \label{corol:impossible_fairness}
    Consider \gmmfour{}, where every minority point participates in the ACA by flipping all \(y{=}0\) labels to \(y{=}1\).
    Then, under sufficiently separable clusters, with high probability, the EqOd of the ERM classifier minimizing the logistic loss will asymptotically approach $0.5$.
\end{proposition}

The formal Proposition~\ref{prop-full-impossible_fairness}, which holds for a broader family of distributions, is provided in \cref{ap:unfair_proof} along with all necessary assumptions. It can be extended to any dimensionality \(\reals^{d}\) using techniques similar to those in \citet{chaudhuriWhyDoesThrowing2023}.
Although Proposition~\ref{corol:impossible_fairness} is not a formal lower bound, it emphasizes an important limitation: ACA restricted to the minority cannot generally achieve zero EqOd violation, even under very advantageous conditions involving a maximum-sized collective, a strong strategy, and a disregard for accuracy.
This limitation stands in contrast to standard firm-side bias mitigation methods, which can in principle achieve zero EqOd violation.
There are several reasons why relabeling alone may not be enough to achieve zero EqOd violation.
For one, relabeling according to the counterfactual implicitly assumes that the label is determined by the same features across the majority and minority, but this assumption may fail under distribution shift between the groups.
To illustrate, consider a firm training a classifier to screen candidates for a managerial position.
Majority members may be more educated, while minority members may have more hands-on experience rather than formal education.
In this case, a counterfactual label associated with the majority is disjoint from the features associated with the minority, rendering the signal erasure strategy irrelevant.
Future work could study this problem and determine when it is beneficial to change the features as well.

\begin{figure*}
    \centering
    \begin{subfigure}{0.23\linewidth}
        \centering
        \caption{\compas{}}
        \includegraphics[width=\linewidth]{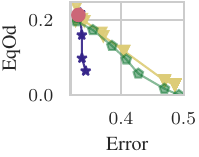}
    \end{subfigure}
    \hfill
    \begin{subfigure}{0.23\linewidth}
        \centering
        \caption{\adult{}}
        \includegraphics[width=\linewidth]{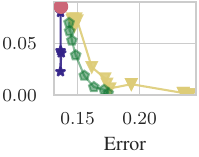}
    \end{subfigure}
    \hfill
    \begin{subfigure}{0.23\linewidth}
        \centering
        \caption{\hsls{}}
        \includegraphics[width=\linewidth]{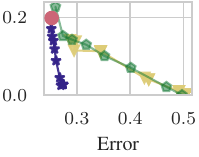}
    \end{subfigure}
    \hfill
    \begin{subfigure}{0.23\linewidth}
        \centering
        \caption{\acsincome{}}
        \includegraphics[width=\linewidth]{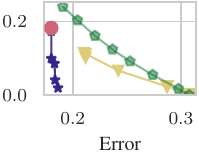}
    \end{subfigure}
    \includegraphics{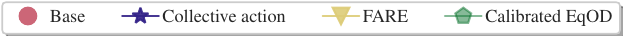}

    \caption{
    The user-side method cannot achieve zero EqOd violation, while the firm-side pre-processing method FARE~\cite{jovanovicFAREProvablyFair2023} and the post-processing method calibrated equalized odds~\cite{pleissFairnessCalibration2017} attain 0 EqOd with large error.
    However, \byprobability{} has lower EqOd violation than the base classifier, with a smaller error than the firm-side methods.
    }
    \label{fig:eqod_zero}
\end{figure*}

We empirically corroborate the findings of Proposition~\ref{corol:impossible_fairness} on real-world datasets by examining the EqOd--accuracy tradeoff of several fair ML methods.
\cref{fig:eqod_zero} compares the Pareto fronts of \byprobability{} with established firm-side methods.
We observe that the lowest EqOd violation achievable by \byprobability{} is greater than that of the firm-side approaches.
However, the firm-side methods arrive at zero EqOd violation only at the cost of prohibitively high prediction error.
In the region where the error is small compared to the base classifier, the EqOd violation of \byprobability{} is comparable to that of the firm-side methods.

\paragraph{Estimating counterfactuals.}
The methods from \cref{sec:method} estimate which individuals would receive a counterfactual label that is different from their original label.
However, the success lower bound in \cref{eq:success_bound} assumes perfect knowledge of \basedist{} and of the underlying causal model.
To account for estimation error, we model the collective's prediction as the output of an algorithm \(\mathcal{A}\left( x\right)\approx\arg\max_{y}\basedist\left(y|x_{A\leftarrow0}\right)\) with error rate \lblerror{}, defined as
\begin{equation} \label{eq:lbl_error}
\lblerror:=\basedist (\mathcal{A}\left(x\right)\neq\arg\max_{y^{\prime}}\basedist\left[y^{\prime}|g\left(x\right)\right]) .
\end{equation}
Given this definition, we derive the following lower bound on success, proved in \cref{ap:thm1_proof}.
\begin{proposition}%
    \label{prop:success_label_error}
For an algorithm \(\mathcal{A}\left( x\right)\) with label error \(\lblerror<1/2\), the success of the collective is bounded by
    \begin{equation} \label{eq:error_success_bound}
        S\left(\alpha\right)\geq1-\frac{2\left(1-\alpha\right)}{\left(1-2\lblerror\right)\alpha}\tau-\frac{\epsilon}{\left(1-\epsilon\right)\left(1-2\lblerror\right)\alpha}.
    \end{equation}
\end{proposition}
This bound recovers \cref{eq:success_bound} when \(\lblerror{=}0\), but higher \lblerror{} worsens the bound.
Next, we show how to use FRL to reduce the error \lblerror{}, thereby improving the lower bound.

\begin{figure*}
    \centering
    \begin{subfigure}{0.23\linewidth}
        \centering
        \caption{\compas{}}
        \includegraphics[width=\linewidth]{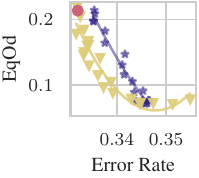}
        \label{fig:knn_sub_compas}
    \end{subfigure}
    \hfill
    \begin{subfigure}{0.23\linewidth}
        \centering
        \caption{\adult{}}
        \includegraphics[width=\linewidth]{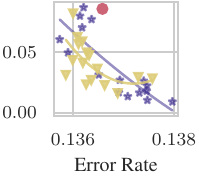}
        \label{fig:knn_sub_adult}
    \end{subfigure}
    \hfill
    \begin{subfigure}{0.23\linewidth}
        \centering
        \caption{\acsincome{}}
        \includegraphics[width=\linewidth]{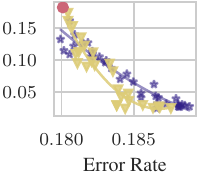}
        \label{fig:knn_sub_income}
    \end{subfigure}
    \hfill
    \begin{subfigure}{0.23\linewidth}
        \centering
        \caption{\birdsfull{}}
        \includegraphics[width=\linewidth]{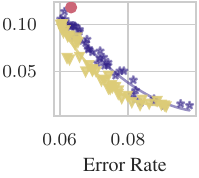}
        \label{fig:knn_sub_birds}
    \end{subfigure}
    \includegraphics{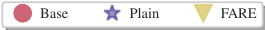}

    \caption{
    The Pareto fronts for using a fair representation when computing the KNN for \bydistance{} dominate the Pareto fronts for KNN computed on untransformed features.
    The blue stars represent the KNN without transforming the data, and the yellow triangles represent the KNN when the data is transformed using FARE~\cite{jovanovicFAREProvablyFair2023}.
    The lines are fitted to guide the eye.
    }
    \label{fig:compare_rep}%
\end{figure*}

\paragraph{Impact of feature representations}
Since the methods \bylabel{} and \bydistance{} rely on KNN, they are sensitive to the choice of distance metric and feature representation.
In \cref{chapter:results}, we used Euclidean distance in the original feature space, which is convenient but could be suboptimal.
Here, we explore whether FRL can learn a more suitable representation space for KNN.
A \emph{fair representation} maps the data into a space where the group-based bias is removed, while preserving informative features.
Intuitively, such representations may help \bylabel{} and \bydistance{} to better estimate the counterfactual labels.
To formalize this intuition, we consider predicting the counterfactual label of minority points using a $1$-NN classifier on majority data, i.e., assigning each minority point the label of its nearest neighbor in the majority.
In settings where the minority and majority are differently distributed~(e.g.,~\gmmfour{}), this task can be challenging.
The following informal result compares the error of $1$-NN in the original feature space to its error in FRL.%
\begin{proposition}[Informal]%
    \label{prop:corlblerror}
     Let data be drawn from \gmmfour{}, and \(\lblerror_{\text{plain}}\) denote the error of a $1$-NN classifier that assigns the label of the nearest majority neighbor in the original feature space.
     Then there exists a fair representation in which a $1$-NN classifier achieves error \(\lblerror_{\text{FRL}}\) such that, asymptotically with respect to the dataset size, $\lblerror_{\text{FRL}} \leq \lblerror_{\text{plain}}$.
\end{proposition}
The formal statement, \cref{thm:rep_lbl_err}, suggests that FRL can reduce the counterfactual label error~\lblerror{} of KNN methods, raising the lower bound of the success according to \cref{prop:success_label_error}.
Empirically, \cref{fig:compare_rep} indicates that applying FARE~\citep{jovanovicFAREProvablyFair2023} before the KNN step improves the Pareto front for \bydistance{}.
Yet, methods that rely purely on predictive information, such as \byprobability{}, can perform worse, due to inadvertent removal of features.

    \section{Related Work} \label{sec:related_work}
Fairness in ML often comes at the cost of reduced classification accuracy, leading to the well-documented accuracy-fairness tradeoff~\citep{menonCostFairnessBinary2018,zhaoInherentTradeoffsLearning2022,dehdashtianUtilityfairnessTradeoffsHow2024,sadeghiCharacterizingTradeoffInvariant2022}.
In response, previous work has proposed fairness interventions at different stages of the ML pipeline: pre-processing methods modify the training data before learning~\citep{kamiranClassifyingDiscriminating2009,luongKNNImplementationSituation2011,zemelLearningFairRepresentations2013,jovanovicFAREProvablyFair2023}, in-processing methods adjust the learning algorithm itself~\citep{agarwalReductionsApproachFair2018,namLearningFailureDebiasing2020,sagawaDistributionallyRobustNeural2020,liuJustTrainTwice2021}, and post-processing methods correct the predictions of a trained (unfair) classifier~\citep{hardtEqualityOpportunitySupervised2016,alghamdiAdultCOMPASFair2022,tifreaFRAPPEGroupFairness2024,cruzUnprocessingSevenYears2024}.
A firm can introduce any of these categories into its pipeline, while users, who control only their data, can only partially implement pre-processing methods.
However, as mentioned in \cref{ssec:collective}, feature-changing pre-processing methods such as fair representation learning~\citep{zemelLearningFairRepresentations2013,jovanovicFAREProvablyFair2023} demand changing features at inference time.

Still, some pre-processing methods change only the labels, similarly to our proposed collective action.
The method by \citet{luongKNNImplementationSituation2011} compares minority KNN and majority KNN and flips labels according to the difference in positive labels between the two groups of neighbors.
This method resembles \bylabel{}, with the difference that \bylabel{} examines only the majority KNN in order to approximate the counterfactual.
The approach of \citet{kamiranClassifyingDiscriminating2009} trains a regressor to predict \(y=1\) outcome probabilities, flips minority members with \(y=0\) labels and high predicted probabilities to \(y=1\), and similarly flips majority \(y=1\) labels to \(y=0\).
Flipping from both groups is intended to preserve the error of the classifier.
Our method \byprobability{} differs by training the regressor only on the majority to better approximate the counterfactuals.
Since this approach requires flipping the labels of majority members as well, it cannot be completely adopted by the collective.
In \cref{ssec:compare_prior} we compare \byprobability{} to CND and KDP, and find that our method is more efficient in terms of the number of label flips.
However, as \cref{fig:compare_flips} demonstrates, the datasets exhibit different sensitivities to the counterfactual approximation method, and it is not guaranteed that our proposed methods will always perform better.
This uncertainty is a result of several factors, including the effect of the data representation (as illustrated in \cref{fig:compare_rep}) and the fact that the collective modifies only the labels, even though the counterfactual features may also be different.

    \section{Conclusion}
This work demonstrates that user-side methods can effectively reduce unfairness in machine learning.
While much of the existing fairness research has focused on firm-side methods, these often come at a cost that may not be worth it to the firm.
This tradeoff emphasizes the importance of studying user-side approaches for bias mitigation.
We show empirically that ACA can considerably reduce unfairness on a variety of datasets, though not completely.
Importantly, we also examine the limitations of collectives composed only of minority members, and how success is affected by approximating the counterfactual labels.
Our proposed methods require collective members to relabel themselves, which often comes at a price, as users have to report labels that differ from their uncoordinated labels.
However, real-world studies show that minority members can willingly participate in collective action to benefit their demographic after being encouraged by their community~\citep{begenyPowerIngroupPromoting2022} or by the media~\citep{saleemWhenHowNegative2021}, and can be efficiently mobilized at scale~\citep{mcadamPoliticalProcessDevelopment1999,michelsonMeetingChallengeLatino2005}.
Overall, this paper shows a practical use case of collective action in the hopes of sparking further research into applications of ACA and user-side methods for social good.

{}

\section*{Acknowledgements}
OB is supported by the International Max Planck Research School for Intelligent Systems (IMPRS-IS).
OB conducted this work at the university of Copenhagen and was funded by ELSA --- European Lighthouse on Secure and Safe AI funded by the European Union under Grant Agreement No. 101070617.  AS acknowledges the Novo Nordisk Foundation for support via the Startup grant (NNF24OC0087820) and VILLUM FONDEN via the Young Investigator program (72069).

\bibliographystyle{unsrtnat}
\bibliography{references}

\pagebreak{}
\section*{Appendix}
\appendix
\crefalias{section}{appendix}
\crefalias{subsection}{appendix}
\crefalias{subsubsection}{appendix}

\section{Motivating examples} \label{ap:motivation}
To recognize real-world problems that lend themselves well to collective action for fairness, one needs to look for the following few characteristics:
\begin{itemize}
    \item Firm and goal: A firm trains a predictive model primarily to minimize average error, with little incentive to protect minority groups.
    \item End-users: People who use the platform and whose behavior generates data for the firm's dataset.
    \item Disadvantaged group: A subgroup of end-users who are treated unfairly.
    \item Relabeling possibility: How the minority can relabel themselves to make the trained classifier fairer.
\end{itemize}

Here we suggest several cases that satisfy these characteristics.
\begin{enumerate}
    \item Content moderation
    \begin{itemize}
        \item Firm and goal: A global social-media company optimizes a high-recall harmful-content detector measured on its largest user pools.
        \item End-users: Everyday users of the platform who can flag offensive content.
        \item Disadvantaged group: Slurs, insults, or cultural references specific to minority communities are not flagged often enough, so the model fails to detect harmful content in those groups' languages.
        \item Relabeling possibility: The minority flags borderline content from their community that the platform's global guidelines ignore.
    \end{itemize}
    \item Resume screening
    \begin{itemize}
        \item Firm and goal: A multinational HR firm trains a classifier to extract skills from resumes.
        \item End-users: Job applicants submitting resumes.
        \item Disadvantaged group: Applicants from a disadvantaged minority may lack formal education and degrees compared to the majority, but may have informal training which the classifier ignores.
        \item Relabeling possibility: Applicants can reframe their work experience, e.g. framing working at a store as being a salesperson, or managing shifts as managerial experience.
    \end{itemize}
    \item Medical treatment prediction
    \begin{itemize}
        \item Firm and goal: A nationwide insurer builds a treatment-recommendation model to minimize average costs and adverse events.
        \item End-users: Patients who report their treatment outcomes (pain levels, recovery time, side effects).
        \item Disadvantaged group: Minority groups may experience different side effects or recovery rates from the majority, so the model recommends suboptimal treatments for them.
        \item Relabeling possibility: Individual patients record more detailed outcomes rather than underreporting, e.g., consistently marking ``still in pain'' instead of ``fine''.
    \end{itemize}
    \item Credit scoring
    \begin{itemize}
        \item Firm and goal: A lender trains a credit-risk model to predict defaults and set loan terms, using historical repayment data.
        \item End-users: Borrowers whose repayment or default becomes training labels.
        \item Disadvantaged group: Disadvantaged groups may not have credit cards or have never taken loans, and only deal with cash but still pay their bills. These actions are ``credit-invisible''.
        \item Relabeling possibility: A borrower can report their paid bills, such as rent or utilities, as repaid loans. These become additional positive repayment labels.
    \end{itemize}
    \item Recommender systems
    \begin{itemize}
        \item Firm and goal: A streaming platform trains a recommender system to maximize engagement, heavily weighted toward mainstream content~\cite{baumannAlgorithmicCollectiveAction2024}.
        \item End-users: Users who like, skip, or re-listen to songs.
        \item Disadvantaged group: Niche genres or local musicians get suppressed, as engagement data mostly comes from the majority’s preferences.
        \item Relabeling possibility: Users can promote underrepresented content by repeatedly listening, liking, or playlisting it.
    \end{itemize}

\end{enumerate}

\section{Preliminaries}

\subsection{Statistical parity and equalized odds} \label{ap:fairness_metrics}
Among the various ways fairness can be defined in machine learning, group fairness is one of the most studied.
Group fairness requires that a model's predictions should not systematically differ between protected groups.
One standard measure of this is statistical parity (SP), which captures the difference in the probability of a positive prediction across groups.
Formally, it is defined as
\begin{equation} \label{eq:SP}
    \text{SP}\left(\classifier\right)=\left|P\left[h\left(x\right)=1|a=1\right]-P\left[h\left(x\right)=1|a=0\right]\right|,
\end{equation}
where a smaller SP value indicates fairer treatment across groups.
However, SP does not account for the ground-truth labels \(y\), and thus optimizing for SP can degrade the overall accuracy.
For example, a classifier that always predicts \(\hat{y}=1\) will have perfect SP but a high prediction error.
Alternatively, a stricter notion called equalized odds (EqOd)~\cite{hardtEqualityOpportunitySupervised2016} requires that both the true positive rate and false positive rate be equal across groups.
Here the EqOd difference is defined as

{\small\begin{equation} %
\text{EqOd}\left(\classifier\right)=\frac{1}{2}\sum_{z=0,1}\left|P\left[h\left(x\right)=1|a=1,y=z\right]-P\left[h\left(x\right)=1|a=0,y=z\right]\right|.
\end{equation}}

\subsection{Suboptimal Bayes classifier}
\label{ap:clf-opt}
\begin{definition}[$\epsilon$-suboptimal classifier]
\label{def:clf-opt}
{
A classifier $f:\mathcal{X}\to\mathcal{Y}$ is $\epsilon$-suboptimal on a set $\mathcal{X}^{\prime}\subseteq\mathcal{X}$ under the distribution $\PP$ if there exists a $\PP^\prime$ with $\mathrm{TV}\left(\PP_{Y|X=x},\PP^\prime_{Y|X=x}\right)\leq \epsilon$  such that for all $x\in \mathcal{X}^{\prime}$
\begin{equation*}
\label{eq:bayes-opt}
f\left(x\right) = \argmax_{y\in\mathcal{Y}} \PP^\prime\left(y|x\right).
\end{equation*}
}
\end{definition}
$\mathrm{TV}\left(\cdot,\cdot\right)$ is the total variation distance between two distributions.
The definition is discussed further in \citet{hardtAlgorithmicCollectiveAction2023}.

\section{Theoretical Results and Proofs} \label{ap:proofs}

\subsection{Counterfactual fairness as success} \label{ap:corl1_proof}

\begin{proposition}[Formal version of Proposition~\ref{prop:success_cf_inf}] 
    \label{prop:success_cf}
Let \(h_\alpha(x)=\arg\max_{y}
P_\alpha(Y=y\mid X=x)\) be a deterministic Bayes classifier, and let
\(g(x)=x_{A\leftarrow 0}\). Suppose that \(g(x)=x\) for majority
individuals with \(A=0\), and let \(\beta=P_0(A=1)>0\). Then
\(S(\alpha)=1 \) if and only if \[
h_\alpha(X)=h_\alpha(X_{A\leftarrow 0})
\quad P_0\text{-a.s. conditional on }A=1.\]
\end{proposition}

\begin{proof}
By definition of the success \(S(\alpha)\) we have,
\[
S(\alpha)=P_0[h_\alpha(X)=h_\alpha(g(X))].
\]
For majority individuals, \(A=0\), the intervention \(A\leftarrow 0\) leaves the group
membership unchanged, so \(g(X)=X\). Hence the event
\(h_\alpha(X)=h_\alpha(g(X))\) holds with probability one conditional on \(A=0\).
Therefore,
\[
S(\alpha)
=
1-\beta
+
\beta P_0[
h_\alpha(X)=h_\alpha(X_{A\leftarrow 0})
\mid A=1
].
\]
Since \(\beta>0\), we have \(S(\alpha)=1\) if and only if
\[
P_0[
h_\alpha(X)=h_\alpha(X_{A\leftarrow 0})
\mid A=1
]=1,
\]
which is exactly the claimed almost-sure counterfactual invariance for minority
individuals.
\end{proof}

\subsection{Impossibility of fairness under ERM}
\label{ap:unfair_proof}
The following proposition follows the structure of Theorem 6 in \citet{chaudhuriWhyDoesThrowing2023}. For a vector \(x\in\reals^d\), let \(D\left(x\right)\) denote a distribution on \(\reals^d\) with mean \(x\). Let \(p\) and \(m\) be the number of majority and minority samples, respectively, with \(p\gg m\). 

\begin{assumption}[Concentration Condition, Assumption 2 from \citet{chaudhuriWhyDoesThrowing2023}]
\label{assm:concentration_interval}
Let $x_1, \ldots, x_n\overset{\text{i.i.d.}}{
\sim} D(0)$ in \(\reals^d\). There exist  maps $X_{\max},c,C:\mathbb{Z}_+\times [0,1]\times \mathbb{Z}_+ \rightarrow \mathbb{R}$ such that for all $n\geq n_0$,  all $\delta \in (0,1)$, and all unit vectors $v\in\reals^{d}$, with probability at least $1 - \delta$
\[ \max_{i\in\{1\cdots,n\}}\left\{ v^{\top}x_{i}\right\} \in\big[X_{\max}(n,\delta,d)-c(n,\delta,d),X_{\max}(n,\delta,d)+C(n,\delta,d)\big] \]    
and $\lim_{n\rightarrow \infty }C(n,\delta,d) =0$, $\lim_{n\rightarrow \infty}c(n,\delta,d) =0$. 
\end{assumption}

\paragraph{Data Model} Labels \(y\in\{-1,1\}\) and protected attribute \(a\in\{-1,1\}\) define four groups whose class-conditional distributions share the same shape \(D(\cdot)\) but have different means:
\[
x\mid (y,a) \sim D\!\big(y\mu + y a \psi\big),
\]
where \(\mu,\psi\in \reals^{2}\) and \(\mu\perp \psi\), with \(\hat{\mu}=\mu/\norm{\mu}\) and \(\hat{\psi}=\psi/\norm{\psi}\). For concreteness, take \(\mu=\norm{\mu}(0,1)^{\top}\) and \(\psi=\norm{\psi}(1,0)^{\top}\). Without loss of generality, let the majority attribute be \(a_M=+1\) (the minority is \(a_m=-1\)). Thus the two majority means lie on the positive diagonal \(\pm(\mu+\psi)\) and the two minority means on the negative diagonal \(\pm(\mu-\psi)\).

Let \(A^M,B^M,A^m,B^m\) denote centered samples from \(D(0)\), where
\(A\) indexes examples with original label \(+1\), and \(B\) indexes examples
with original label \(-1\). We analyze the SVM objective in signed coordinates.
Thus, before relabeling, majority signed examples are centered at
\(\mu+\psi\), while minority signed examples are centered at \(\mu-\psi\).

The majority examples are not relabeled. For the minority examples, write
\[
A^m=A^{m,+}\cup A^{m,-},
\qquad
B^m=B^{m,+}\cup B^{m,-},
\]
where the superscript denotes the training label after relabeling. Hence
\(A^{m,+}\) and \(B^{m,-}\) keep their original labels, while
\(A^{m,-}\) and \(B^{m,+}\) are flipped.   

If the minority were absent, the ERM SVM converges to the spurious direction
\[
w_{\text{spu}}^{\mathrm{maj}}\ \propto\ \mu+\psi.
\]

\begin{proposition}
\label{prop-full-impossible_fairness}
Suppose $D(0)$ satisfies \cref{assm:concentration_interval} and

{\small\begin{equation}\label{eq:size-gap}
X_{\text{max}}\left(p,\delta,2\right)-X_{\text{max}}\left(m,\delta,2\right)\geq
2\max\{\norm{\mu},\norm{\psi}\}
+
c\left(p,\delta,2\right)
+
C\left(m,\delta,2\right).
\end{equation}}

Then, for any (possibly adversarial) relabeling of the minority (which preserves separability) of training examples, if $p\rightarrow\infty$, with probability at least $1-4\delta$, gradient descent of ERM with logistic loss converges to the same spurious solution $w_{\text{spu}}^{*}\propto \mu+\psi$.
Under a centrally symmetric $D(0)$ and when \(\|\mu\|=\|\psi\|\), this limit satisfies $\text{EqOd}\left(w_{\text{spu}}^{*}\right)\to 0.5$.
\end{proposition}

\begin{proof}

    As in \citet{chaudhuriWhyDoesThrowing2023}, the limiting direction
of gradient descent on the logistic loss over separable data is the
hard-margin SVM direction. It is therefore enough to characterize the
SVM direction.

Let
\[
w_{\alpha,\sigma}
=
\alpha\hat{\mu}
+
\sigma\gamma\hat{\psi},
\qquad
\gamma=\sqrt{1-\alpha^2},
\qquad
\alpha\in[-1,1],
\qquad
\sigma\in\{-1,+1\}.
\]
We write \(\gamma\) instead of \(\beta\) to avoid confusing this
coefficient with a population fraction. For fixed \((\alpha,\sigma)\),
define the terms
\[
\begin{aligned}
M_A^\sigma(\alpha)
&:=
\sup_{u\in A^M} -w_{\alpha,\sigma}^{\top}u,
&
M_B^\sigma(\alpha)
&:=
\sup_{u\in B^M} w_{\alpha,\sigma}^{\top}u,
\\
R_{A,+}^\sigma(\alpha)
&:=
\sup_{u\in A^{m,+}} -w_{\alpha,\sigma}^{\top}u,
&
R_{A,-}^\sigma(\alpha)
&:=
\sup_{u\in A^{m,-}} w_{\alpha,\sigma}^{\top}u,
\\
R_{B,+}^\sigma(\alpha)
&:=
\sup_{u\in B^{m,+}} -w_{\alpha,\sigma}^{\top}u,
&
R_{B,-}^\sigma(\alpha)
&:=
\sup_{u\in B^{m,-}} w_{\alpha,\sigma}^{\top}u.
\end{aligned}
\]
We use the convention that the supremum over an empty relabeled subset is
\(-\infty\).

The negative-label side of the SVM objective contains the majority negative
examples, the flipped \(A^{m,-}\) examples, and the kept \(B^{m,-}\) examples.
The positive-label side contains the majority positive examples, the kept
\(A^{m,+}\) examples, and the flipped \(B^{m,+}\) examples. Therefore the SVM
objective can be written as
\[
\begin{aligned}
F_\sigma(\alpha)
=
&
\max\Big\{
M_B^\sigma(\alpha)-\alpha\norm{\mu}-\sigma\gamma\norm{\psi},
\\
&\hspace{1.7cm}
R_{A,-}^\sigma(\alpha)+\alpha\norm{\mu}-\sigma\gamma\norm{\psi},
\\
&\hspace{1.7cm}
R_{B,-}^\sigma(\alpha)-\alpha\norm{\mu}+\sigma\gamma\norm{\psi}
\Big\}
\\
&+
\max\Big\{
M_A^\sigma(\alpha)-\alpha\norm{\mu}-\sigma\gamma\norm{\psi},
\\
&\hspace{1.7cm}
R_{A,+}^\sigma(\alpha)-\alpha\norm{\mu}+\sigma\gamma\norm{\psi},
\\
&\hspace{1.7cm}
R_{B,+}^\sigma(\alpha)+\alpha\norm{\mu}-\sigma\gamma\norm{\psi}
\Big\}.
\end{aligned}
\]

By \cref{assm:concentration_interval}, for the majority samples of size \(p\),
there exist \(X_p,c_p,C_p\) such that, with probability at least \(1-4\delta\),
uniformly over \(\alpha\) and \(\sigma\),
\[
M_A^\sigma(\alpha),\,M_B^\sigma(\alpha)
\in
[X_p-c_p,\ X_p+C_p].
\]
For the minority samples, every relabeled subset is contained in the original
minority sample, so
\[
R_{A,+}^\sigma(\alpha),
R_{A,-}^\sigma(\alpha),
R_{B,+}^\sigma(\alpha),
R_{B,-}^\sigma(\alpha)
\le
X_m+C_m,
\]
where
\[
X_m:=X_{\max}(m,\delta,2),
\quad
C_m:=C(m,\delta,2).
\]
Using \cref{eq:size-gap}, we obtain, uniformly over \(\alpha\) and \(\sigma\),
\[
M_A^\sigma(\alpha)-R_{\cdot}^{\sigma}(\alpha)
\ge
2\max\{\norm{\mu},\norm{\psi}\},
\quad
M_B^\sigma(\alpha)-R_{\cdot}^{\sigma}(\alpha)
\ge
2\max\{\norm{\mu},\norm{\psi}\},
\]

for every term \(R_{\cdot}^{\sigma}\) appearing above.

We now show that the majority entries attain both inner maxima. Consider the
first maximum. Comparing the majority term with the flipped \(A^{m,-}\) term,
\[
\begin{aligned}
&
\left(
M_B^\sigma(\alpha)-\alpha\norm{\mu}-\sigma\gamma\norm{\psi}
\right)
-
\left(
R_{A,-}^\sigma(\alpha)+\alpha\norm{\mu}-\sigma\gamma\norm{\psi}
\right)
\\
&\qquad
=
M_B^\sigma(\alpha)-R_{A,-}^\sigma(\alpha)
-
2\alpha\norm{\mu}
\\
&\qquad
\ge
2\norm{\mu}-2\alpha\norm{\mu}
\ge
0.
\end{aligned}
\]
Comparing the majority term with the kept \(B^{m,-}\) term,
\[
\begin{aligned}
&
\left(
M_B^\sigma(\alpha)-\alpha\norm{\mu}-\sigma\gamma\norm{\psi}
\right)
-
\left(
R_{B,-}^\sigma(\alpha)-\alpha\norm{\mu}+\sigma\gamma\norm{\psi}
\right)
\\
&\qquad
=
M_B^\sigma(\alpha)-R_{B,-}^\sigma(\alpha)
-
2\sigma\gamma\norm{\psi}
\\
&\qquad
\ge
2\norm{\psi}-2\gamma\norm{\psi}
\ge
0.
\end{aligned}
\]

Thus the first maximum is attained by the majority term. The same comparisons
show that the second maximum is also attained by the majority term. Hence
\[
F_\sigma(\alpha)
=
M_A^\sigma(\alpha)
+
M_B^\sigma(\alpha)
-
2\alpha\norm{\mu}
-
2\sigma\gamma\norm{\psi}.
\]

Write
\[
E_\sigma(\alpha)
:=
M_A^\sigma(\alpha)+M_B^\sigma(\alpha)-2X_p.
\]
The concentration condition gives
\[
E_\sigma(\alpha)\in[-2c_p,2C_p]
\]
uniformly over \(\alpha\) and \(\sigma\). Therefore
\[
F_+(\alpha)
=
2X_p+E_+(\alpha)
-
2\big(\alpha\norm{\mu}+\gamma\norm{\psi}\big),
\]
whereas
\[
F_-(\alpha)
=
2X_p+E_-(\alpha)
-
2\big(\alpha\norm{\mu}-\gamma\norm{\psi}\big).
\]

For the \(\sigma=+1\) branch, \(F_+(\alpha)\) is minimized at
\[
\alpha_g
=
\frac{\norm{\mu}}
{\sqrt{\norm{\mu}^2+\norm{\psi}^2}},
\]
By~\cref{lem:chaudhuri_approxmax1}, the minimizer of \(F_+\) converges
to \(\alpha_g\) as \(p\to\infty\).
\end{proof}

 \begin{lemma}[Approximate Maximization Lemma - I, Lemma 14 from \citet{chaudhuriWhyDoesThrowing2023}]
\label{lem:chaudhuri_approxmax1}
    Let $F(\alpha) = f(\alpha) + g(\alpha)$ where $g(\alpha) = \alpha u + \sqrt{1 - \alpha^2} v$, $u, v > 0$, and $f(\alpha)\in [-L, U]$.
    Let $\alpha_F \in\argmax_\alpha F(\alpha)$, and let $\alpha_g = \frac{u}{\sqrt{u^2 + v^2}}\in\argmax_{\alpha}g(\alpha)$. 

    Then, the angle between $(\alpha_F, \sqrt{1 - \alpha_F^2})$ and $(\alpha_g, \sqrt{1 - \alpha_g^2})$ is at most $\cos^{-1}\left( 1 - \frac{L + U}{\sqrt{u^2 + v^2}}\right)$, and $\max_\alpha F(\alpha)\geq \sqrt{u^2 + v^2} - L$. 

\end{lemma}

For the \(\sigma=-1\) branch, minimizing \(F_-(\alpha)\) is
asymptotically equivalent to maximizing
\[
\alpha\norm{\mu}-\sqrt{1-\alpha^2}\norm{\psi}.
\]
This quantity is at most \(\norm{\mu}\). By contrast,
\[
\max_{\alpha}
\left[
\alpha\norm{\mu}
+
\sqrt{1-\alpha^2}\norm{\psi}
\right]
=
\sqrt{\norm{\mu}^2+\norm{\psi}^2}
>
\norm{\mu}
\]
whenever \(\psi\neq 0\). Since \(c_p,C_p\to 0\), the \(\sigma=+1\) branch
has the strictly smaller asymptotic SVM objective. Hence the optimal sign is
\(\sigma=+1\). Therefore,
\[
w^{*}
\longrightarrow
\alpha_g\hat{\mu}
+
\gamma_g\hat{\psi}
=
\frac{\mu+\psi}
{\sqrt{\norm{\mu}^2+\norm{\psi}^2}}
\propto
\mu+\psi.
\]
Thus, independently of how the minority samples were relabeled in training,
the limiting ERM direction is the same majority spurious direction
\[
w_{\text{spu}}^{*}\propto \mu+\psi.
\]

Under a centrally symmetric \(D(0)\) and if \(\norm{\mu}=\norm{\psi}\), the
majority group \((a=+1)\) is separated in the limiting spurious direction,
while the minority signed means lie along \(\mu-\psi\). Since
\[
(\mu+\psi)^{\top}(\mu-\psi)
=
\norm{\mu}^{2}-\norm{\psi}^{2}
=
0,
\]
the limiting classifier cuts the two minority classes symmetrically. Hence
\[
\mathrm{TPR}_{a=+1}\to1,
\quad
\mathrm{FPR}_{a=+1}\to0,
\quad
\mathrm{TPR}_{a=-1}\to\frac12,
\quad
\mathrm{FPR}_{a=-1}\to\frac12.
\]
Consequently,
\[
\mathrm{EqOd}(w_{\text{spu}}^{*})\to0.5.
\]

Finally, since the data are linearly separable under the stated assumptions,
gradient descent on the logistic loss converges in direction to the
hard-margin SVM solution
\citep{soudryImplicitBiasGradient2018,jiImplicitBiasGradient2019}. Hence the
same limiting direction holds for ERM with logistic loss.

This result can also be extended to \(\reals^{d}\) using techniques similar to those in \citet{chaudhuriWhyDoesThrowing2023}.
This result also encompasses the 4-Gaussian mixture model \gmmfour{} used in Section~\ref{ssec:minority_restrictions} as a special case, leading to Proposition~\ref{corol:impossible_fairness}.

\subsection{Success Bound With Label Error} 
\label{ap:thm1_proof}
The following proof uses Lemma 11 from \citet{hardtAlgorithmicCollectiveAction2023}.
\begin{lemma}[Lemma 11 from \citet{hardtAlgorithmicCollectiveAction2023}]
\label{lemma:hardt_coupling}
Suppose that $P,P'$ are two distributions such that $\mathrm{TV}(P,P')\leq \epsilon$. Take any two events $E_1,E_2$ measurable under $P,P'$. If $P(E_1) > P(E_2) + \frac{\epsilon}{1-\epsilon}$, then $P'(E_1) > P'(E_2)$.
\end{lemma}

Restatement of \cref{prop:success_label_error}:
\begin{proposition}%
For an algorithm \(\mathcal{A}\left( x\right)\) with label error \(\lblerror<1/2\), the success of the collective is bounded by
    \begin{equation}
        S\left(\alpha\right)\geq1-\frac{2\left(1-\alpha\right)}{\left(1-2\lblerror\right)\alpha}\tau-\frac{\epsilon}{\left(1-\epsilon\right)\left(1-2\lblerror\right)\alpha}.
    \end{equation}
\end{proposition}

\begin{proof}
    
This proof follows closely the proof of Theorem 5 by \citet{hardtAlgorithmicCollectiveAction2023}.
We start under the assumption of an optimal Bayes classifier, setting \(\epsilon=0\).

When the new label \(y^{\prime}\) is wrong with probability \lblerror{}, then we can think of the collective as a union of two sub-collectives: one with the correct label and one with the incorrect label.
In the binary case this can be formulated with correct subcollective \(P^{+}\) as having label \(y^{\prime}=\arg\max_{y}P_{0}\left(y|g\left(x\right)\right)\) and the incorrect subcollective \(P^{-}\) as having label \(y^{\prime}=\arg\min_{y}P_{0}\left(y|g\left(x\right)\right)\).
Then we can write the train distribution as 
\begin{equation}
    \begin{aligned}
    P_{\alpha} & =\alpha\left(\lblerror P^{-}+\left(1-\lblerror\right)P^{+}\right)+\left(1-\alpha\right)P_{0}\\
 & =\alpha\lblerror P^{-}+\left(1-\lblerror\right)\alpha P^{+}+\left(1-\alpha\right)P_{0}.
\end{aligned}
\end{equation}

Denote \(y^{*}\left(x\right)=\arg\max_{y}P_{0}\left(y|g\left(x\right)\right)\), then the probability to get prediction \(y^{*}\) is
\begin{equation}
    \begin{aligned}P_{\alpha}\left(y^{*}|x\right) & =\alpha\lblerror P^{-}\left(y^{*}|x\right)+\left(1-\lblerror\right)\alpha P^{+}\left(y^{*}|x\right)+\left(1-\alpha\right)P_{0}\left(y^{*}|x\right)\\
 & =\left(1-\lblerror\right)\alpha+\left(1-\alpha\right)P_{0}\left(y^{*}|x\right),
\end{aligned}
\end{equation}
and the probability to get the prediction \(y\neq y^{*}\) is
\begin{equation}
    \begin{aligned}P_{\alpha}\left(y|x\right) & =\alpha\lblerror P^{-}\left(y|x\right)+\left(1-\lblerror\right)\alpha P^{+}\left(y|x\right)+\left(1-\alpha\right)P_{0}\left(y|x\right)\\
 & =\alpha\lblerror+\left(1-\alpha\right)P_{0}\left(y|x\right),
\end{aligned}
\end{equation}
where \(P^{+}\left(y^{*}|x\right)=1\), \(P^{-}\left(y^{*}|x\right)=0\), \(P^{+}\left(y|x\right)=0\), \(P^{-}\left(y|x\right)=1\) by definition.

A Bayes classifier \(\classifier\) returns the most probable label \(\classifier\left(x\right)=\arg\max_{y}P\left(y|x\right)\).
Therefore, a Bayes classifier will output \(y^{*}\) if the probability is greater, which can be written as the condition
\begin{equation} \label{eq:bayes_conditions}
    \begin{aligned}P_{\alpha}\left(y^{*}|x\right) & >P_{\alpha}\left(y|x\right)\\
\left(1-\lblerror\right)\alpha+\left(1-\alpha\right)P_{0}\left(y^{*}|x\right) & >\alpha\lblerror+\left(1-\alpha\right)P_{0}\left(y|x\right)\\
\left(1-2\lblerror\right)\alpha & >\left(1-\alpha\right)\left(P_{0}\left(y|x\right)-P_{0}\left(y^{*}|x\right)\right).
\end{aligned}
\end{equation}
Let \(\tau\left(x\right)=\max_{y}\left[P_{0}\left(y|x\right)-P_{0}\left(y|g\left(x\right)\right)\right]\), then
\begin{equation}
    \begin{aligned}P_{0}\left(y|x\right)-P_{0}\left(y^{*}|x\right) & \leq P_{0}\left(y|x\right)-P_{0}\left(y|g\left(x\right)\right)+P_{0}\left(y^{*}|g\left(x\right)\right)-P_{0}\left(y^{*}|x\right)\\
 & \leq2\tau\left(x\right).
\end{aligned}
\end{equation}
With that, the condition in \cref{eq:bayes_conditions} can be written as
\begin{equation} \label{eq:tau_condition}
    \left(1-2\lblerror\right)\alpha>2\left(1-\alpha\right)\tau\left(x\right).
\end{equation}

With that, the success can be bounded as
\begin{equation}
    \begin{aligned}S & =P_{0}\left[f\left(x\right)=f\left(g\left(x\right)\right)\right]\\
 & =P_{0}\left[f\left(x\right)=y^{*}\left(x\right)\right]\\
 & \geq P_{0}\left[\left(1-2\lblerror\right)\alpha>2\left(1-\alpha\right)\tau\left(x\right)\right]\\
 & =P_{0}\left[1-\frac{2\left(1-\alpha\right)}{\left(1-2\lblerror\right)\alpha}\tau\left(x\right)>0\right]\\
 & =\mathbb{E}_{x\sim P_{0}}\left[\boldsymbol{1}\left\{ 1-\frac{2\left(1-\alpha\right)}{\left(1-2\lblerror\right)\alpha}\tau\left(x\right)>0\right\} \right]\\
 & \geq\mathbb{E}_{x\sim P_{0}}\left[1-\frac{2\left(1-\alpha\right)}{\left(1-2\lblerror\right)\alpha}\tau\left(x\right)\right]\\
 & =1-\frac{2\left(1-\alpha\right)}{\left(1-2\lblerror\right)\alpha}\tau
\end{aligned}
\end{equation}

\paragraph{With sub-optimality \(\epsilon > 0\)} A consequence of \cref{lemma:hardt_coupling} is to write the condition in \cref{eq:tau_condition} as
\begin{equation}
    \left(1-2\lblerror\right)\alpha>2\left(1-\alpha\right)\tau\left(x\right)+\frac{\epsilon}{1-\epsilon},
\end{equation}
which by following the same steps as with \(\epsilon=0\) results in the final bound
\begin{equation}
    S\left(\alpha\right)\geq1-\frac{2\left(1-\alpha\right)}{\left(1-2\lblerror\right)\alpha}\tau-\frac{\epsilon}{\left(1-\epsilon\right)\left(1-2\lblerror\right)\alpha}.
\end{equation}
\end{proof}

\subsection{Label Error With Better Representation}
\label{ap:lbl_error_proof}

\newcommand{\I}{\mathbb{I}}
\newcommand{\m}[1]{\mu^{#1}}
\newcommand{\s}[1]{\sigma^{#1}}
\newcommand{\Dn}{\mathcal{D}_n}
\newcommand{\R}{\mathbb{R}}
\newcommand{\E}{\mathbb{E}}
\newcommand{\var}{\text{Var}}
\newcommand{\snr}{SNR}
\newcommand{\snrproj}{SNR_\text{proj}}
\newcommand{\mus}{\mu_s}
\newcommand{\muc}{\mu_c}
\newcommand{\muz}{\mu_Z}
\newcommand{\varz}{\sigma_Z^2}
\newcommand{\mutest}{\mu_\text{min}}
\newcommand{\covtest}{\Sigma_\text{min}}
\newcommand{\vproj}{\bar{v}}
\newcommand{\muproj}{\bar{\mu}}
\newcommand{\mutestproj}{\bar{\mu}_\text{min}}
\newcommand{\covtestproj}{\bar{\Sigma}_\text{min}}
\newcommand{\xs}{x_s}
\newcommand{\xc}{x_c}
\newcommand{\Xminus}{X_-}
\newcommand{\Xplus}{X_+}
\newcommand{\sigmas}{\sigma_s}
\newcommand{\sigmac}{\sigma_c}
\newcommand{\N}{\mathcal{N}}
\newcommand{\cdf}{\Phi}
\newcommand{\test}{\text{min}}
\newcommand{\pplus}{p_+}
\newcommand{\pminus}{p_-}
\newcommand{\id}{\mathbbm{1}}
\newcommand{\yhat}{\hat{y}_\text{1NN}}
\newcommand{\yhatn}{\hat{y}_\text{1NN}^{(n)}}
\newcommand{\err}{\text{err}_\text{1NN}}
\newcommand{\mean}[2]{(#1, #2)^\top}
\newcommand{\cov}[2]{\diag(#1, #2)}

In the following we assume a similar setting as in \cref{ap:unfair_proof}, visualized as a 2D distribution in \cref{fig:example_dist}.
We are given the majority data, and tasked with labeling the minority data.
Assume all labels are distributed equally \(\PP[Y=1]=\PP[Y=-1]=\frac{1}{2}\).
The minority features \Xtest{} are distributed as \(\Xtest\sim\N(y\mutest,\covtest)\) with \(\Xtest\in\R^d\).
The label \(\yhatn\) is predicted according to a 1NN classifier from \(n\) majority samples \(\Dn={(x_i, y_i)}_{i=0}^n\).
Majority samples with \(y=+1\) are distributed as \(\Xplus\sim\N(\mu, \Sigma)\), and with \(y=-1\) are distributed as \(\Xminus\sim \N(-\mu, \Sigma)\).

\begin{theorem} \label{thm:rep_lbl_err}
    Assume that $\mutest^\top \Sigma^{-1}\mu < 0$.
    Further, consider the setting with $\covtest = I$, and the minority (i.e., test) distribution introduced above with $\PP[Y=1]=\PP[Y=-1]=0.5$ and $\Xtest\sim\N(y\mutest, \covtest)$.
    
    Then, there exists a projection $P\in\R^{d\times d}$ such that asymptotically for $n\to\infty$, $\err^\text{rep} < \err^\text{raw}$.
\end{theorem}
\begin{proof}

    Consider the projection on the hyperplane perpendicular to $w$, where $w=\frac{\mu-\mutest}{2}$. The assumption $\mutest^\top\Sigma^{-1}\mu<0$ implies $\mu\neq\mutest$, so $w\neq0$ and the projection matrix is well-defined as $P=I-\frac{ww^\top}{w^\top w}$.
    Since $P$ projects onto the subspace orthogonal to $\mu-\mutest$, it satisfies
    \begin{equation}
        P\mu=P\mutest.
    \end{equation}

    We apply \Cref{lemma:err} to obtain closed forms for the asymptotic error of 1NN applied to the initial representation and to the features after the projection $P$. Namely, using the notation $v:=\Sigma^{-1}\mu$ we have:

    \begin{align}
        \err&=\frac{1}{2}\PP_{\Xtest|y=1}[\yhat= -1] + \frac{1}{2}\PP_{\Xtest|y=-1}[\yhat=1]\\
        &=\frac{1}{2} \left( 1-\cdf\left( \frac{v^\top\mutest}{\sqrt{v^\top\covtest v}} \right) \right) + \frac{1}{2} \cdf\left( \frac{-v^\top\mutest}{\sqrt{v^\top\covtest v}} \right) \\
        &=1-\cdf\left( \frac{v^\top\mutest}{\sqrt{v^\top\covtest v}} \right)\\
        &= 1 - \cdf(\snr),
    \end{align}

    where we used the fact that $\cdf(-z)=1-\cdf(z)$ and we denote $\snr:=\frac{v^\top\mutest}{\sqrt{v^\top\covtest v}}$.

    Since the numerator is negative by assumption and the denominator is positive, $\snr<0$, and therefore $\err^\text{raw}>1/2$.

    We now analyze the projected representation using coordinates on the image of $P$.
    Let $U\in\R^{d\times r}$ be an orthonormal basis for $\operatorname{Im}(P)$, where $r=\operatorname{rank}(P)$, so $U^\top U=I_r$ and $UU^\top=P$.
    The projected majority distributions in coordinates $z=U^\top x$ are
    \begin{equation}
        Z_+\sim\N(\eta,\Sigma_U),\qquad Z_-\sim\N(-\eta,\Sigma_U),
    \end{equation}
    where $\eta:=U^\top\mu$ and $\Sigma_U:=U^\top\Sigma U$.
    The projected minority distribution is
    \begin{equation}
        Z_{\mathrm{min}}\mid y\sim\N(y\eta,I_r),
    \end{equation}
    because $U^\top\mutest=U^\top\mu$ and $\covtest=I$.

    If $\eta=0$, the projected representation contains no label signal and the balanced-label error is $\err^\text{rep}=1/2$, which is strictly smaller than $\err^\text{raw}$.
    Otherwise, applying \cref{lemma:err} in the projected coordinate system gives
    \begin{equation}
        \err^\text{rep}=1-\cdf(\snrproj),
        \qquad
        \snrproj
        =
        \frac{\eta^\top\Sigma_U^{-1}\eta}
        {\sqrt{(\Sigma_U^{-1}\eta)^\top(\Sigma_U^{-1}\eta)}}.
    \end{equation}
    Since $\Sigma_U$ is positive definite and $\eta\neq0$, the numerator is positive and the denominator is positive, so $\snrproj>0$.
    Therefore $\err^\text{rep}<1/2<\err^\text{raw}$.
    
\end{proof}

\begin{lemma}
\label{lemma:err}
    For a unimodal minority distribution $\Xtest\sim\N(\mutest,\covtest)$ it holds that:
$$\lim_{n\to\infty}\PP_{\Xtest}[\yhatn=-1]=1-\cdf\left( \frac{v^\top\mutest}{\sqrt{v^\top\covtest v}} \right),$$
where $v:=\Sigma^{-1}\mu\text{ and }\cdf\text{ is the CDF of a standard Gaussian}.$
\end{lemma}

\begin{proof}

Let us denote $\yhat:=\lim_{n\to\infty}\yhatn$ and let $\pplus$ and $\pminus$ be the densities of two class-conditional distribution. Notice that the two class conditional training distributions are supported on the entire domain of $\R^d$. Therefore, in the asymptotic regime, the label $\yhat$ at a test point $x$ is given according to the class-conditional distribution that has higher density. Namely, we have:

$$
\yhat = \begin{cases}
-1 & \text{if } \pplus(x) < \pminus(x), \\
1 & \text{otherwise.}
\end{cases}
$$

Given $\Xtest \sim \N(\mutest,\covtest)$, we can then write the probability of predicting $\yhat=-1$ as:
$$\PP_{\Xtest}[\yhat= -1]=\PP_{\Xtest}\left[ \pplus(x) < \pminus(x) \right].$$

Using the closed forms for the pdf of a Gaussian, we write the corresponding log-probabilities as follows:
$$\log \pplus(x) = -\frac{1}{2} (x-\mu)^\top \Sigma^{-1} (x-\mu) + \text{const}.$$

$$\log \pminus(x) = -\frac{1}{2} (x+\mu)^\top \Sigma^{-1} (x+\mu) + \text{const}.$$

Using the fact that $\log$ is monotonically increasing and $\Sigma$ (and by extension $\Sigma^{-1}$) is a symmetric matrix, we can write after some simple calculations:

$$\PP_{\Xtest}[\yhat= -1]=\PP_{\Xtest}[\mu^\top\Sigma^{-1}x<0].$$

Let us denote the random variable $Z:=(\Sigma^{-1}\mu)^\top X$. Since $Z$ is a linear transformation of Gaussian random variable, it is itself Gaussian and we can write its mean and variance as follows:

$$\muz:=v^\top\mutest\text{, and }\varz:=v^\top\covtest v\text{, where }v:=\Sigma^{-1}\mu.$$

After this change of variable, we can rewrite the probability of predicting $\yhat=-1$ as:

\begin{align*}
    \PP_{\Xtest}[\yhat= -1]&=\PP_Z\left[Z<0\right]\\
&=\cdf\left(\frac{0-\E[Z]}{\sqrt{\var[Z]}}\right)\\
&=\cdf\left( \frac{ -v^\top\mutest }{ \sqrt{ v^\top\covtest v } } \right)\\
&=1- \cdf\left( \frac{v^\top\mutest }{ \sqrt{ v^\top\covtest v } } \right).
\end{align*}

\end{proof}

Note that the error from \cref{thm:rep_lbl_err} is defined the same as \lblerror{} (\cref{eq:lbl_error}).
This leads to \cref{prop:corlblerror}.

\section{Technical Details}

\subsection{Datasets} \label{ap:datasets}

\paragraph{COMPAS}
The Correctional Offender Management Profiling for Alternative Sanctions (COMPAS) dataset contains the data of criminal defendants from the Broward County Sheriff's Office in Florida with the task of predicting the recidivism risk.
The label in this dataset represents whether the person re-offended and the sensitive attribute is the race.
We follow the same data cleaning and pre-processing as \citet{alghamdiAdultCOMPASFair2022}.

\paragraph{Adult}
The Adult dataset~\cite{beckerAdult1996} contains demographic features of US citizens and is tasked with predicting the income level of an individual.
The label represents whether the individual has income higher than \$50,000 and the sensitive attribute we use is the race.
We follow the same data cleaning and pre-processing as \citet{alghamdiAdultCOMPASFair2022}.

\paragraph{HSLS}
The High School Longitudinal Study of 2009 (HSLS)~\cite{jeongFairnessImputationDecision2022} contains details of high-school students across the US and the task is to predict the academic success of the students.
The label represents the exam score and the sensitive attribute is the race.
We follow the same data cleaning and pre-processing as \citet{alghamdiAdultCOMPASFair2022}.

\paragraph{ACS-Income} 
\citet{dingRetiringAdultNew2021} offer different classification tasks derived from US census data.
In our work we used the pre-defined task of predicting the level of income denoted as \textit{ACS-Income}, where the data is already pre-processed.
The label represents whether the individual has income higher than \$50,000 and the sensitive attribute is the race.

\paragraph{Waterbirds}
The waterbirds dataset~\cite{sagawaDistributionallyRobustNeural2020} contains images of landbirds and waterbirds superimposed on either land or water backgrounds, with the task of classifying the image as a landbird or a waterbird.
The label represents the type of bird, and the sensitive attribute is whether the background is land or water.
To obtain the features, we used the output of the penultimate layer of a pre-trained ResNet-18 network from \textit{PyTorch}~\footnote{https://pytorch.org/vision/main/models/generated/torchvision.models.resnet18.html}.
We report the results on those features as \birdsfull{}.
We also performed PCA (using \textit{scikit-learn}) and kept the first 85 principal components which retain about 75\% of the variance, and report the results of these components as \birdspca{}.

\paragraph{CivilComments}
The CivilComments dataset~\cite{borkanNuancedMetricsMeasuring2019} is a collection of text comments found on the internet, with the goal of training a classifier to fairly detect toxicity.
For this paper, we modified the dataset to keep only the comments that include either \textit{christian} or \textit{muslim} (but not both), with a label 0 meaning toxic and 1 meaning safe.
To obtain the features, we used the word embeddings given by Hugging Face's \textit{bert-base-uncased} model\footnote{https://huggingface.co/google-bert/bert-base-uncased}.
We report the results on those features as \civilfull{}.
We also performed PCA (using \textit{scikit-learn}) and kept the first 100 principal components which retain about 75\% of the variance, and report the results of these components as \civilpca{}.

\subsection{Training} \label{ap:training}
All classification experiments were trained with \textit{scikit-learn}'s histogram-based gradient boosting classification tree with the default parameters~\footnote{scikit-learn.org/stable/modules/generated/sklearn.ensemble.\\HistGradientBoostingClassifier.html}.
When there was not a pre-defined test set, we set the train-test split as 80-20 before applying the collective action.

The probabilities for \byprobability{} were inferred by training \textit{scikit-learn}'s histogram-based gradient boosting classification tree on the majority data with the default parameters, and using its \textit{predict\_proba} function.
For LFR~\cite{zemelLearningFairRepresentations2013} we used the implementation in \textit{Holistic AI}'s open-source library~\footnote{https://github.com/holistic-ai/holisticai} with the default parameters.
For FARE~\cite{jovanovicFAREProvablyFair2023} we used the official implementation~\footnote{https://github.com/eth-sri/fare} with hyperparameters \(\gamma=0.85\), \(k=200\) and \(n=100\).
For all distance computations we used the Euclidean \(\ell^{2}\)-norm as \(d\left(v,u\right)=\left\Vert v-u\right\Vert _{2}=\sqrt{\sum_{i}\left(v_{i}-u_{i}\right)^{2}}\).

\section{Additional Results} \label{ap:full_results}
\subsection{Comparison with prior work}
\label{ssec:compare_prior}
\begin{figure}
    \centering
    \begin{subfigure}{\linewidth}
        \centering
        \caption{EqOd}
        \includegraphics[width=\linewidth]{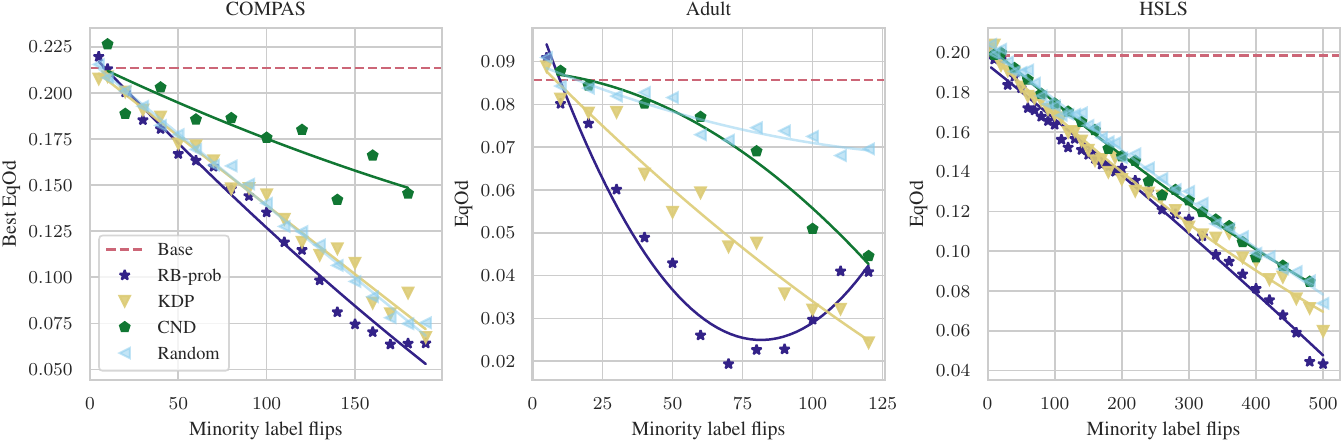}
    \end{subfigure}
    \begin{subfigure}{\linewidth}
        \centering
        \caption{SP}
        \includegraphics[width=\linewidth]{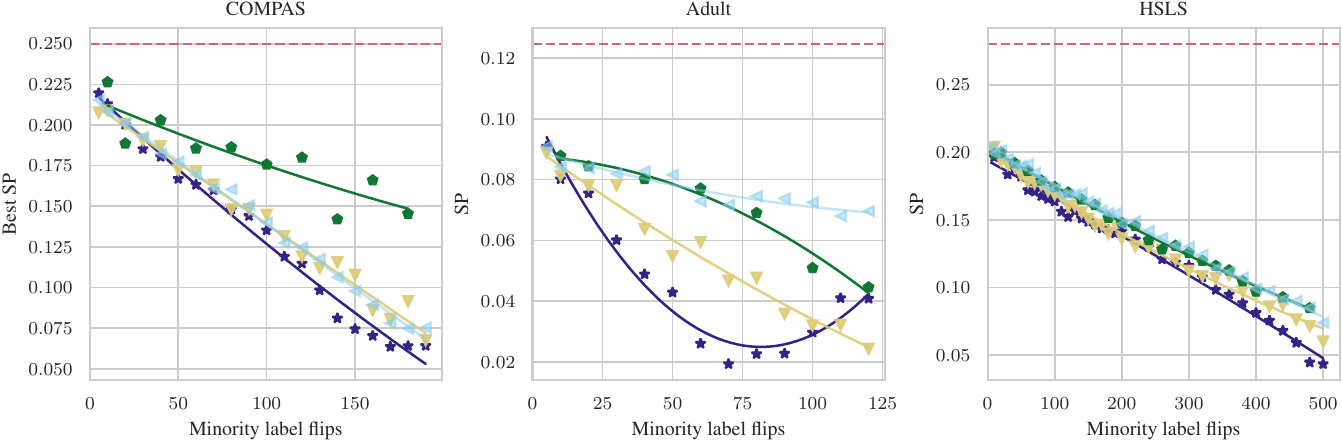}
    \end{subfigure}
    
    \caption{
    EqOd and SP violations per number of label flips for the random baseline, our method \byprobability{}, and the existing methods KDP~\citep{luongKNNImplementationSituation2011} and CND~\citep{kamiranClassifyingDiscriminating2009}.
    Our method is more efficient than prior work, requiring fewer flips to achieve comparable violation levels.
    Note that in this experiment CND could flip any label, while all other methods were restricted to the labels of 30\% of the minority.}
    \label{fig:compare_prior}
\end{figure}

We compare our method \byprobability{} with the existing methods KDP~\cite{luongKNNImplementationSituation2011} and CND~\cite{kamiranClassifyingDiscriminating2009} in \cref{fig:compare_prior}.
Note that CND requires flipping labels for both majority and minority members, and we report the total number of label flips.
\cref{fig:compare_prior} shows that our method, motivated by the counterfactual labeling, is more efficient in terms of required number of label flips than the existing works.

\subsection{Expanded results}
\label{ssec:expanded_results}
The following figures include the results of the experiments reported in the main text using all methods on all datasets, both with EqOd (\cref{eq:EqOd}) and SP (\cref{eq:SP}) as measures of unfairness.

\begin{figure}
    \centering
    \begin{subfigure}{0.46\linewidth}
        \centering
        \caption{EqOd}
        \includegraphics[width=\linewidth]{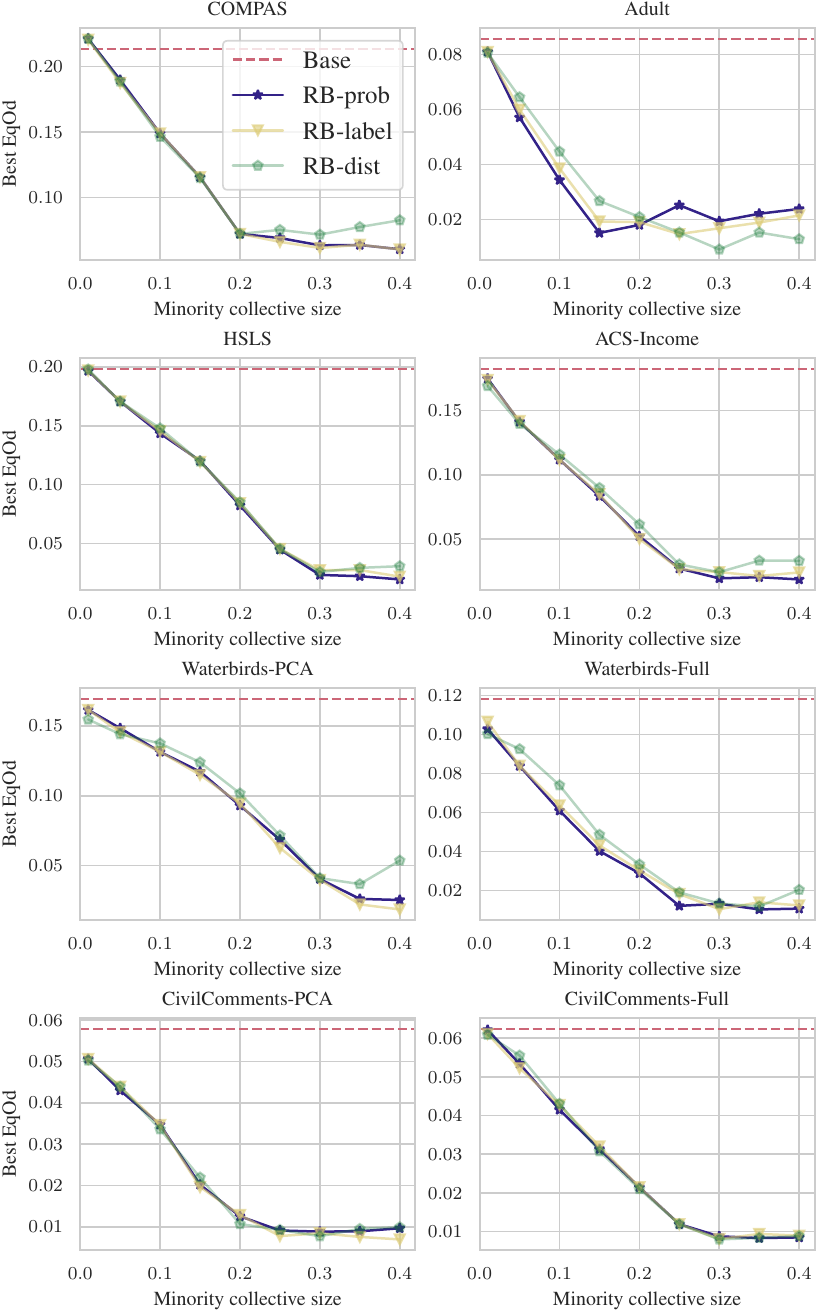}
    \end{subfigure}
    \hfill
    \begin{subfigure}{0.46\linewidth}
        \centering
        \caption{SP}
        \includegraphics[width=\linewidth]{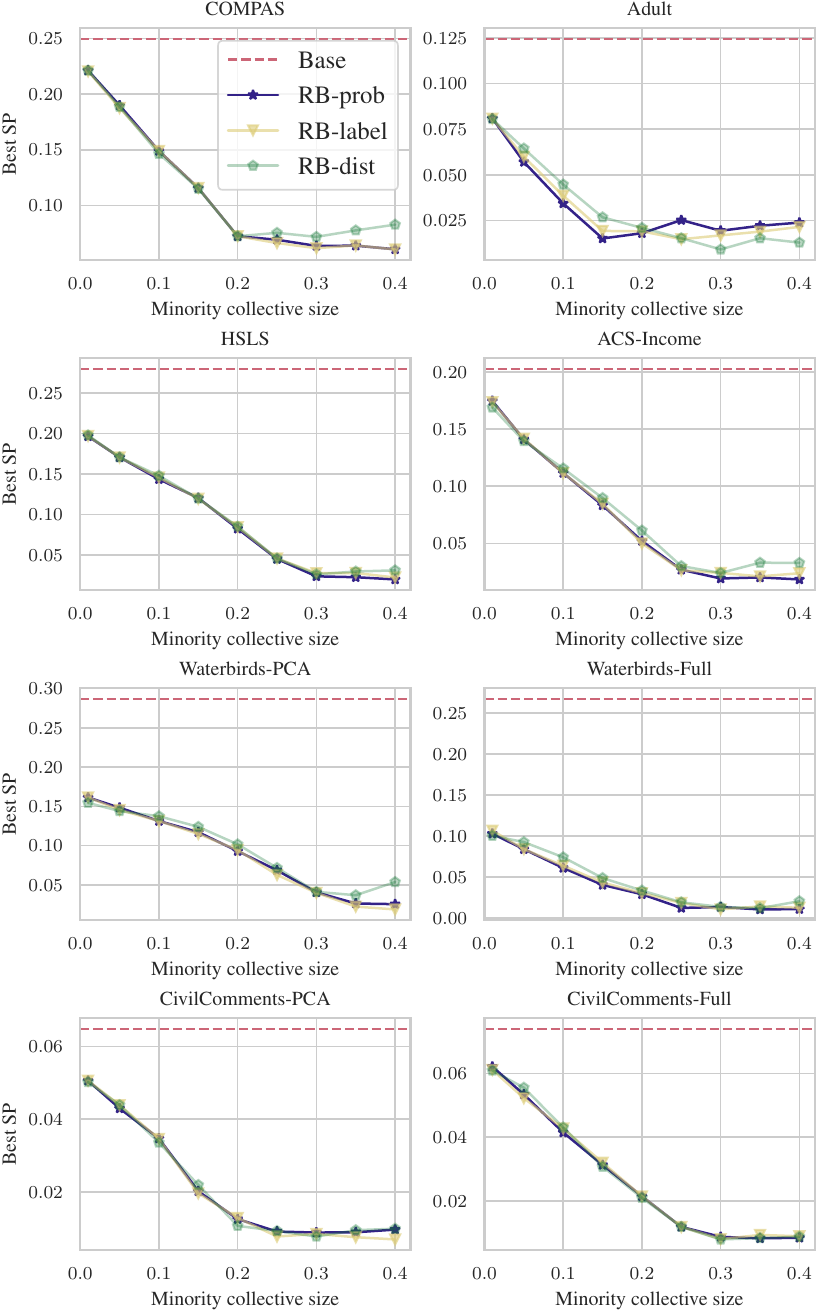}
    \end{subfigure}

    \caption{
    The lowest EqOd and SP violations a collective can achieve decrease as the collective size increases, up to a certain point.
    Each point is a mean of 10 runs, with the standard deviation being smaller than the markers.
    In all the datasets we experimented on, both violation metrics stabilize around \(\strength=0.3\).}
    \label{fig:sp_effect_alpha}
\end{figure}

\begin{figure}
    \centering
    \begin{subfigure}{0.46\linewidth}
        \centering
        \caption{EqOd}
        \includegraphics[width=\linewidth]{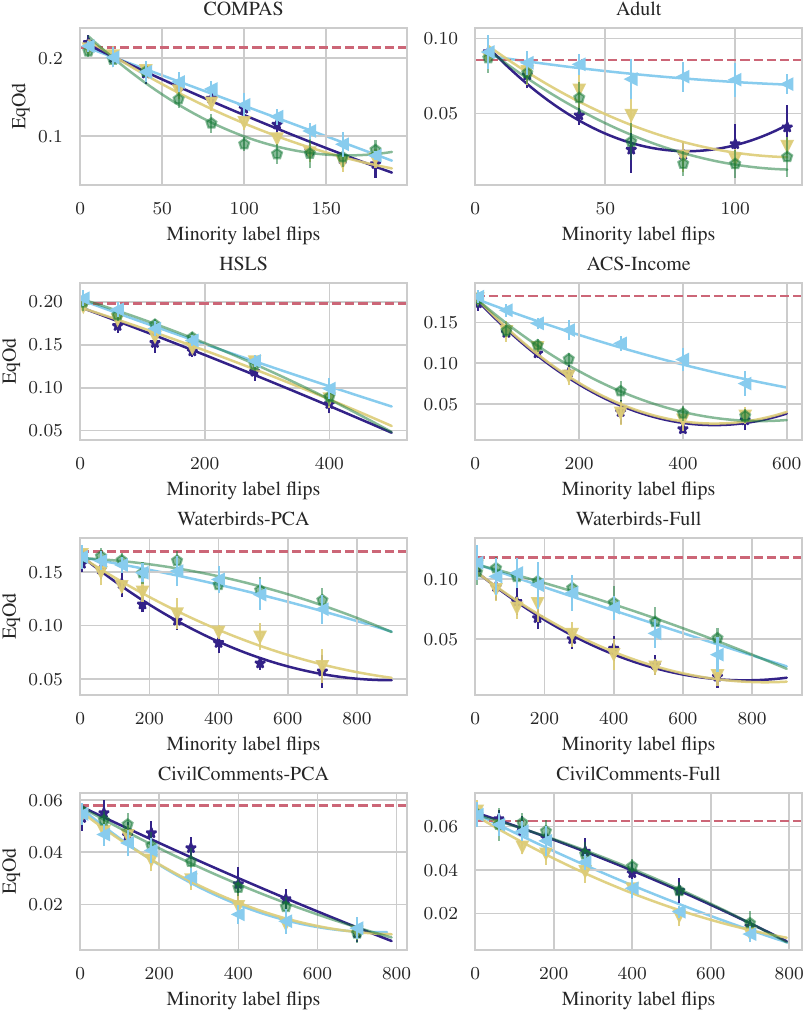}
    \end{subfigure}
    \hfill
    \begin{subfigure}{0.46\linewidth}
        \centering
        \caption{SP}
        \includegraphics[width=\linewidth]{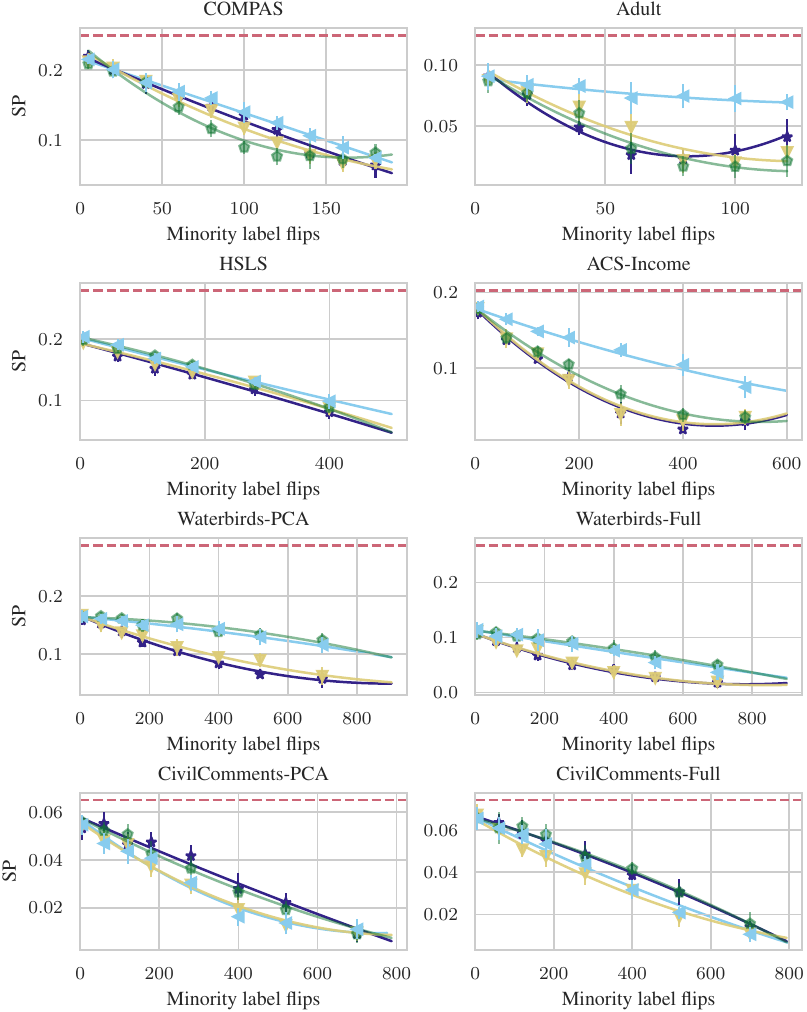}
    \end{subfigure}
    \includegraphics{figures/compare_flips/legend.pdf}
   
    \caption{
    Our proposed methods are consistently more efficient than randomly flipping labels, requiring fewer label flips to attain comparable EqOd or SP violation levels.
    Each marker is the mean of 10 random runs with a specific number of label flips.
    The dashed line shows the corresponding mean violation for a classifier trained on the dataset without collective action.}
    \label{fig:sp_all_flips}
\end{figure}

\begin{figure}
    \centering
    \begin{subfigure}{0.49\linewidth}
        \centering
        \caption{EqOd}
        \includegraphics[width=\linewidth]{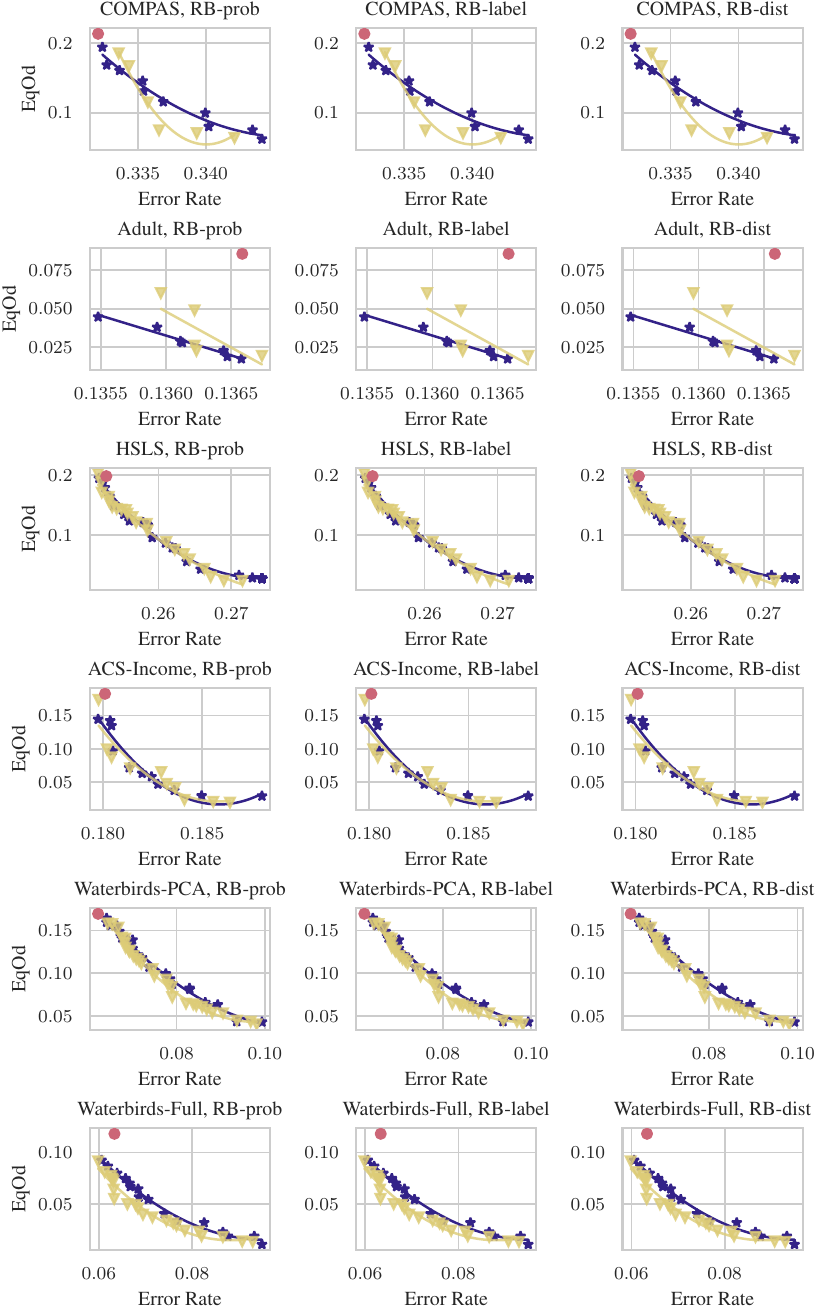}
    \end{subfigure}
    \hfill
    \begin{subfigure}{0.49\linewidth}
        \centering
        \caption{SP}
        \includegraphics[width=\linewidth]{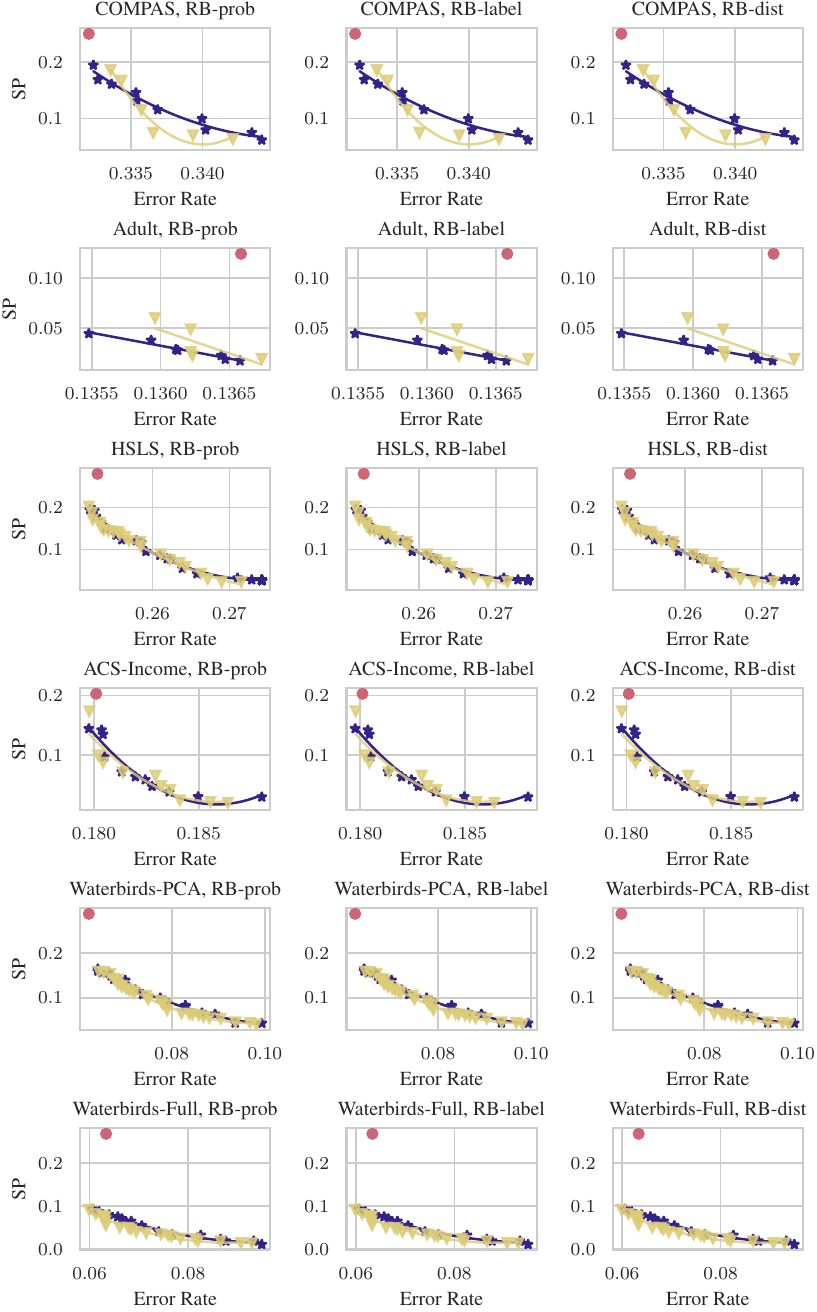}
    \end{subfigure}
    \includegraphics[width=0.98\linewidth]{figures/compare_knowledge/legend.pdf}
    \caption{
    Limiting the knowledge of the collective about the majority does not significantly harm the Pareto front.
    Each point is the mean of 10 runs and the curves are fitted to guide the eye.}
    \label{fig:sp_partial_knowledge}
\end{figure}

\begin{figure}
    \centering
    \begin{subfigure}{0.46\linewidth}
        \centering
        \caption{EqOd} 
        \includegraphics[width=\linewidth]{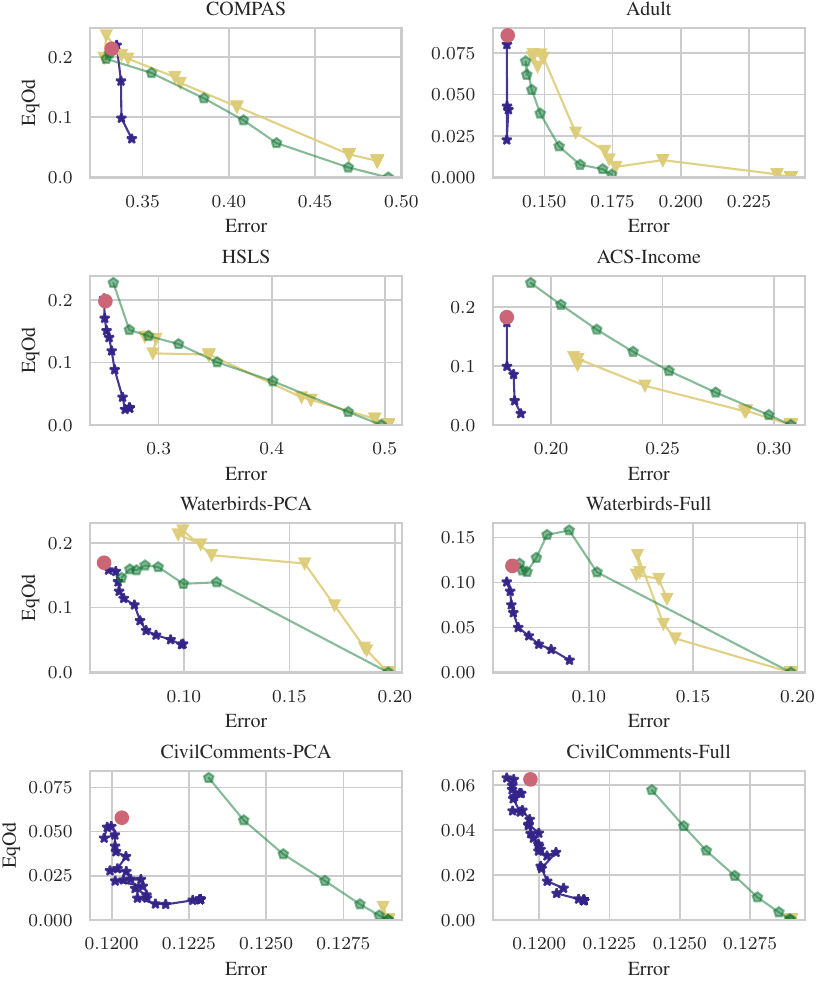}
    \end{subfigure}
    \hfill
    \begin{subfigure}{0.46\linewidth}
        \centering
        \caption{SP} 
        \includegraphics[width=\linewidth]{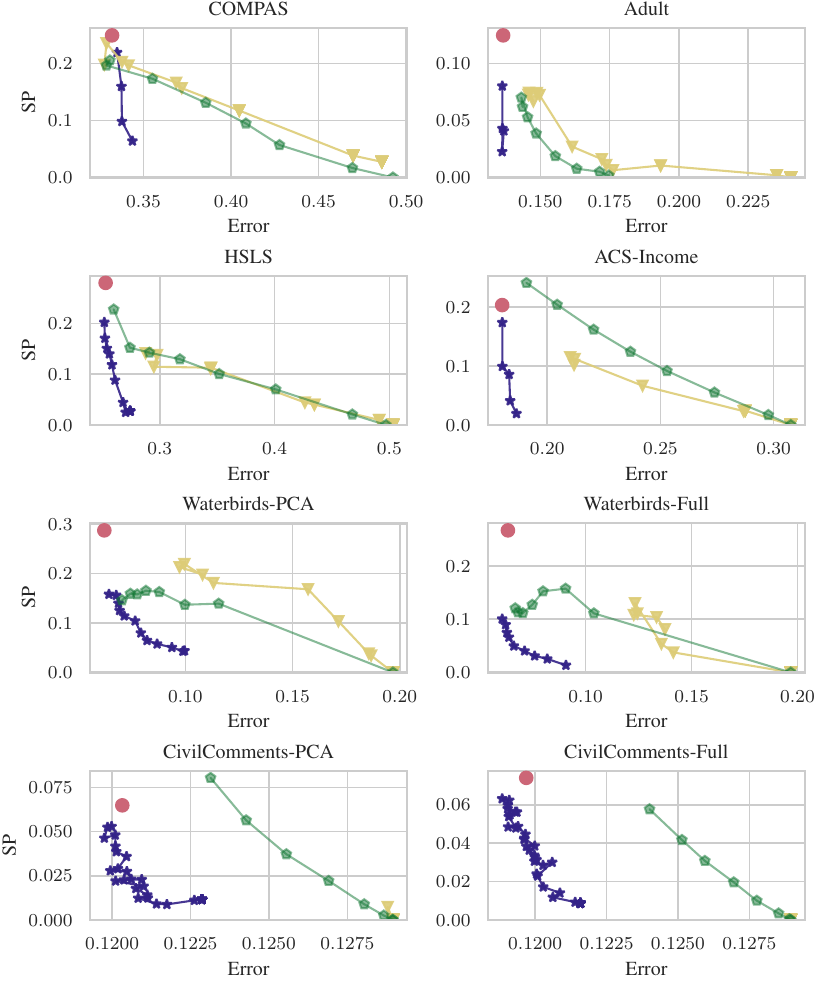}
    \end{subfigure}
    \includegraphics[width=0.98\linewidth]{figures/paretos_processing/legend.pdf}
    \caption{
    The EqOd panel shows that the firm-side pre-processing method FARE~\cite{jovanovicFAREProvablyFair2023} and the post-processing method calibrated equalized odds~\cite{pleissFairnessCalibration2017} attain 0 EqOd with large error.
    Across both EqOd and SP, \byprobability{} with \(\strength=0.3\) (Section~\ref{sec:method}) has much smaller error and lower violations than the base classifier, but does not reach zero violation.}
    \label{fig:sp_all_paretos}
\end{figure}

\begin{figure}
    \centering
    \begin{subfigure}{0.49\linewidth}
        \centering
        \caption{\byprobability{}, EqOd}
        \includegraphics[width=\linewidth]{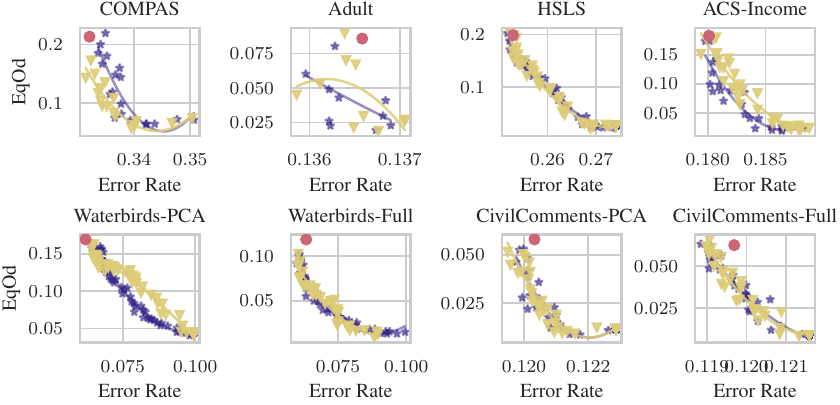}
    \end{subfigure}
    \hfill
    \begin{subfigure}{0.49\linewidth}
        \centering
        \caption{\byprobability{}, SP}
        \includegraphics[width=\linewidth]{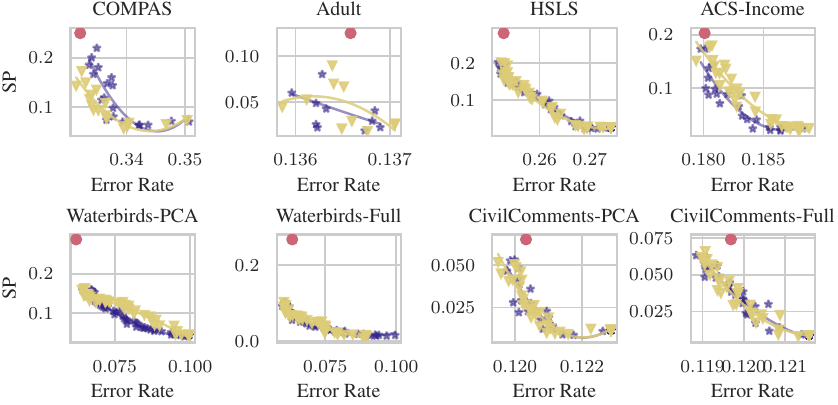}
    \end{subfigure}

    \begin{subfigure}{0.49\linewidth}
        \centering
        \caption{\bylabel{}, EqOd}
        \includegraphics[width=\linewidth]{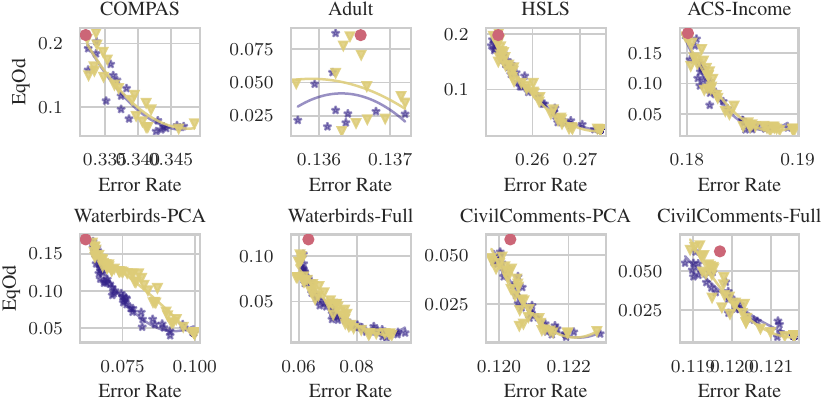}
    \end{subfigure}
    \hfill
    \begin{subfigure}{0.49\linewidth}
        \centering
        \caption{\bylabel{}, SP}
        \includegraphics[width=\linewidth]{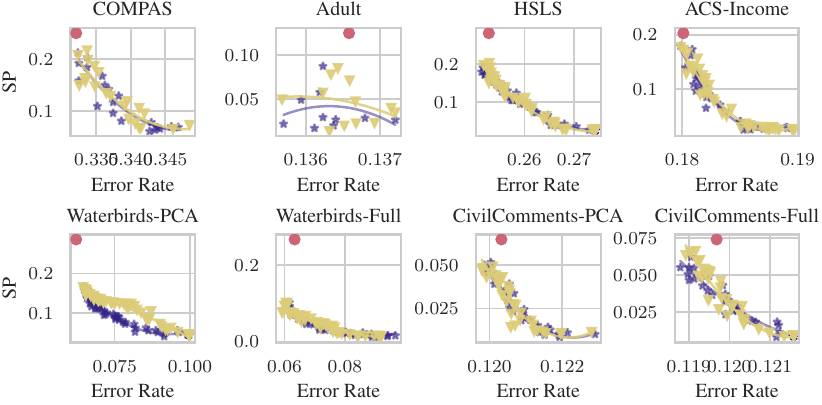}
    \end{subfigure}

    \begin{subfigure}{0.49\linewidth}
        \centering
        \caption{\bydistance{}, EqOd}
        \includegraphics[width=\linewidth]{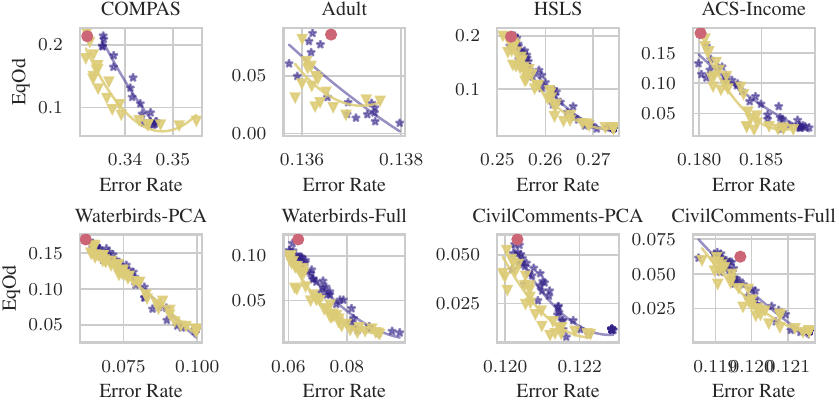}
    \end{subfigure}
    \hfill
    \begin{subfigure}{0.49\linewidth}
        \centering
        \caption{\bydistance{}, SP}
        \includegraphics[width=\linewidth]{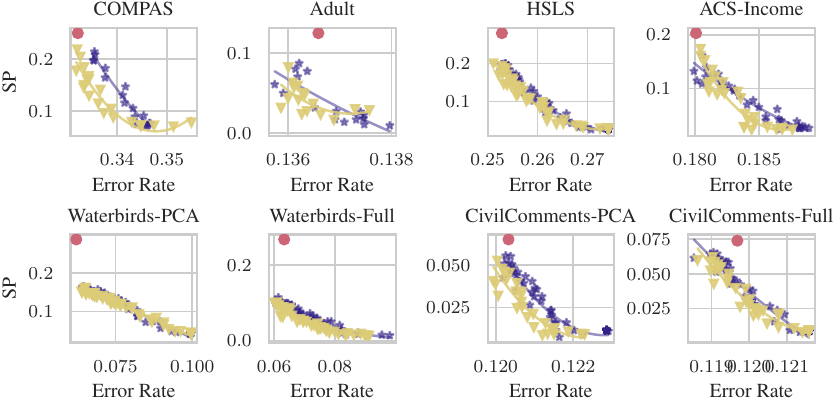}
    \end{subfigure}
    
    \includegraphics{figures/compare_representation/legend.pdf}
    \caption{
    Using a fair representation generally shifts the Pareto fronts lower and left across \byprobability{}, \bylabel{}, and \bydistance{}.
    The blue stars represent each method without transforming the data, and the yellow triangles represent each method after transforming the data using FARE~\citep{jovanovicFAREProvablyFair2023}.
    The lines are fitted by a polynomial of degree 2 to guide the eye.}
    \label{fig:sp_rep_effect}
\end{figure}

\end{document}